# Fast and Sample Efficient Inductive Matrix Completion via Multi-Phase Procrustes Flow

Xiao Zhang[∗†] and Simon S. Du[∗‡] and Quanquan Gu[§]

**Abstract**

We revisit the inductive matrix completion problem that aims to recover a rank-$r$ matrix with ambient dimension $d$ given $n$ features as the side prior information. The goal is to make use of the known $n$ features to reduce sample and computational complexities. We present and analyze a new gradient-based non-convex optimization algorithm that converges to the true underlying matrix at a linear rate with sample complexity only linearly depending on $n$ and logarithmically depending on $d$. To the best of our knowledge, all previous algorithms either have a quadratic dependency on the number of features in sample complexity or a sub-linear computational convergence rate. In addition, we provide experiments on both synthetic and real world data to demonstrate the effectiveness of our proposed algorithm.

## 1 Introduction

Matrix completion method has been used in a wide range of applications such as collaborative filtering for recommendation (Koren et al., 2009), multi-label learning (Cabral et al., 2011) and clustering (Hsieh et al., 2012). In these applications, every entry is modeled as the inner product between factors corresponding to the row and column variables. For example, in movie recommendation, each row factor represents the latent representation of a user and each column factor represents the latent representation of a movie.

In many applications of significant interest, besides the partially observed matrix, side information, in the form of features, is also available. These might correspond to demographic information (genders, occupation) for users or product information (genre, director) in a movie recommender system for example. With such features at hand, one can model an observation as a specific linear interaction between features to reduce the model complexity. Formally, let $\mathbf{L}^* \in \mathbb{R}^{d_1 \times d_2}$ be the unknown low-rank matrix with rank $r$, and let $\mathbf{X}_L \in \mathbb{R}^{d_1 \times n_1}$ and $\mathbf{X}_R \in \mathbb{R}^{d_2 \times n_2}$ be the known feature matrices with $d_1 \geq n_1 \geq r$ and $d_2 \geq n_2 \geq r$. We assume the unknown rank-$r$ matrix $\mathbf{L}^*$ can be represented by $\mathbf{X}_L \mathbf{M}^* \mathbf{X}_R^\top$ for some unknown matrix $\mathbf{M}^* \in \mathbb{R}^{n_1 \times n_2}$. Thus instead of learning a large $d_1 \times d_2$ matrix $\mathbf{L}^*$, we only need to recover a smaller low-rank matrix $\mathbf{M}^*$. This inductive approach has been applied successfully in many applications including collaborative filtering (Abernethy et al., 2009; Menon et al., 2011; Chen et al., 2012), multi-label learning (Xu et al., 2013; Si et al., 2016), semi-supervised clustering (Yi et al., 2013; Si et al., 2016), gene-disease prediction (Natarajan and Dhillon, 2014) and blog recommendation (Shin et al., 2015).

From the theoretical point of view, side information allows us to reduce the overall sample and computational complexities. Xu et al. (2013) and Jain and Dhillon (2013) pioneered the theoretical investigation in this direction. Specifically, Xu et al. (2010) adapted the convex relaxation approach (Candès and Recht, 2009; Candès and Tao, 2010) and requires only $O(rn \log n \log d)$[1] samples to recover the underlying matrix, which we believe is tight up to logarithmic factors. However, the computational cost is usually high because they need to

---

[∗]Equal Contribution
[†]Department of Computer Science, University of Virginia, Charlottesville, VA 22904, USA; e-mail:xz7bc@virginia.edu
[‡]Machine Learning Department, Carnegie Mellon University, Pittsburgh, PA 15213; e-mail: ssdu@cs.cmu.edu
[§]Department of Computer Science, University of Virginia, Charlottesville, VA 22904, USA; e-mail: qg5w@virginia.edu

[1]For the ease of presentation, we assume $d_1 = d_2 = d$ and $n_1 = n_2 = n$ when discussing complexities.

solve a nuclear norm minimization problem, which is inherently slow due to its high per-iteration complexity and non-strongly convex objective function (c.f. Equation (2) in Xu et al. (2013)), which does not have linear convergence rate. On the other hand, Jain and Dhillon (2013) (also see Zhong et al. (2015)) proposed an algorithm which first does a spectral initialization to obtain a coarse estimate, then uses alternating minimization to refine the estimate. Their algorithm has a locally linear rate of convergence but requires $O(r^3 n^2 \log n \log(1/\epsilon))$ samples, which has an unsatisfactory quadratic dependency on $n$ and cannot achieve exact recovery because sample complexity also depends on the target accuracy $\epsilon$. A natural and open question is:

**Can we recover the ground truth matrix at a linear rate with sample complexity linear in $n$?**

In this paper, we answer this question affirmatively. Specifically, we propose a multi-phase gradient-based algorithm that converges to the underlying true matrix at a linear rate with sample complexity linearly depending on $n$ and logarithmically depending on $d$. Our algorithm is a novel and highly nontrivial extension of Procrustes Flow (Tu et al., 2015) in which we add an additional phase to reduce the variance of gradient estimate, and therefore we call it Multi-Phase Procrustes Flow. The main challenges and technical insights are summarized in the following section.

## 1.1 Main Challenges and Technical Insights

In recent years, a surge of non-convex optimization algorithms for estimating low-rank matrices have been established. A typical procedure is first to do a spectral initialization to obtain a coarse estimate, and then to use Burer-Monteiro factorization (Burer and Monteiro, 2003) with projected gradient descent (a.k.a., Procrustes flow) on the partially observed entries to recover the underlying matrix, where the projection is introduced to control the variance of gradient descent (Tu et al., 2015; Zheng and Lafferty, 2016; Yi et al., 2016). Our proposed algorithm also follows this framework. However, direct adaptation does not achieve the desired statistical and computational rates. Statistically, in the classical matrix completion setting, after the initialization phase, the variance of the gradient is at a smaller order than the magnitude of expected gradient for *all* iterations. However, in our setting, because of limited samples, such uniform bound does not hold. Computationally, the projection step in the inductive setting is more costly than that in the classical setting because we need to solve a convex quadratically-constrained-quadratic-programming (QCQP) problem (c.f. Section 4).

Our first key observation is that the variance of the gradient converges to 0 at a faster rate than the magnitude of expectation of the gradient. Therefore, if the iterate is close enough to the optimum, say in a ball with radius $O(1/n)$ around the optimum, the desired uniform bound still holds. Further, this observation also indicates when we are close to the optimum, projection step is *not* needed, i.e., vanilla gradient descent suffices. Nevertheless, with limited samples, the spectral initialization cannot directly achieve this goal.

Our second key observation is that after a rough spectral initialization, if we use fresh samples to calculate the gradient at each iteration, the variance is still small compared with the expectation of the gradient. In light of this, we add a new phase to the original algorithm where we use fresh samples to estimate the gradient at each iteration and use projected gradient descent to refine our estimation. Though the projection is costly, we only need $O(r \log n)$ iterations to converge to a ball with radius $O(1/n)$ around the optimum, since gradient descent in our problem enjoys a linear rate of convergence. Putting all these phases together, we propose the first gradient-based algorithm that requires only $O\left(r^2 n \log n \log d\right)$ samples and converges to the ground truth matrix at a linear rate.

## 1.2 Organization and Notation

The remainder of this paper is organized as follows. The most relevant works are reviewed in Section 2. In Section 3, we introduce necessary background and formally state our problem. We give details of our proposed algorithms in Section 4, In Section 5, we present our main theoretical results. We conduct experiments to compare our method with existing algorithms in Section 6, and conclude in Section 7. We defer our additional experiments and detailed proofs to Appendix.

Capital boldface letters such as $\mathbf{A}$ are used for matrices, and $[\ell]$ is used to denote the index set $\{1, 2, \ldots, \ell\}$. Denote the $d \times d$ identity matrix by $\mathbf{I}_d$. Let $\mathbf{A}_{i,*}$, $\mathbf{A}_{*,j}$ and $A_{ij}$ be the $i$-th row, $j$-th column and $(i,j)$-th entry of matrix $\mathbf{A}$, respectively. Denote the $\ell$-th largest singular value of $\mathbf{A}$ by $\sigma_\ell(\mathbf{A})$ and its projection onto the index set $\Omega$ by $\mathcal{P}_\Omega(\mathbf{A})$, i.e., the $(i,j)$-th entry of $\mathcal{P}_\Omega(\mathbf{A})$ is equal to $A_{ij}$ if $(i,j) \in \Omega$ and zero otherwise. Let $\|\mathbf{x}\|_2$ be the $\ell_2$ norm of a $d$-dimensional vector $\mathbf{x} \in \mathbb{R}^d$. Let $\|\mathbf{A}\|_F$, $\|\mathbf{A}\|_2$ be the Frobenius norm and the spectral norm of matrix $\mathbf{A}$ respectively. The largest $\ell_2$ norm of its rows is defined as $\|\mathbf{A}\|_{2,\infty} = \max_i \|\mathbf{A}_{i,*}\|_2$. For any two sequences $\{a_n\}$ and $\{b_n\}$, we say $a_n = O(b_n)$ if there exists a positive constant $C$ such that $a_n \leq C\, b_n$. All the constants such as $c_0, c_1, c_2, \ldots$ throughout the paper are universal constants.

## 2 Related Work

### 2.1 Low-Rank Matrix Completion

Classical approach for matrix completion relies on convex relaxation (Candès and Recht, 2009; Candès and Tao, 2010; Recht, 2011; Chen, 2015; Allen-Zhu et al., 2017), which can be solved by nuclear norm minimization. Such methods usually have nearly tight sample complexity (Balcan et al., 2017). However, due to the use of nuclear norm and non-strongly convex objective function, they cannot achieve linear convergence rate and often scale cubically with the dimension. Some faster algorithms have been proposed (Jain and Netrapalli, 2015) but they often incur additional sample complexity.

To reduce the runtime complexity, various non-convex algorithms have been proposed. Jain et al. (2013); Hardt (2014); Hardt and Wootters (2014); Gamarnik et al. (2017) showed that with proper initialization, alternating minimization enjoys a linear convergence rate. Proofs of these works often build on a general analytical framework, noisy-power-method (Hardt and Price, 2014; Balcan et al., 2016). Nevertheless, the sample complexity often depends on the inverse of target accuracy. Thus these methods often cannot recover the ground truth matrix exactly.

Another line of research studies the landscape of optimization problem and showed that with proper modification of objective function, all local minima are global and all saddle points are strict (Bhojanapalli et al., 2016b; Ge et al., 2016, 2017). Therefore, perturbed gradient descent algorithms can solve this non-convex problem efficiently (Ge et al., 2015; Jin et al., 2017; Du et al., 2017a). However, to guarantee the landscape having nice properties, they all require the sample complexity scales with the fourth power of the rank, which is suboptimal.

Lastly, Tu et al. (2015); Zhao et al. (2015); Zheng and Lafferty (2015); Sun and Luo (2015); Bhojanapalli et al. (2016a); Zheng and Lafferty (2016); Yi et al. (2016); Wang et al. (2016); Ma et al. (2017) proposed first-order algorithms to solve low-rank matrix estimation problems. Similar to Jain et al. (2013); Hardt and Wootters (2014); Hardt (2014), these algorithms first use spectral initialization to find a good starting point, but then instead of performing alternating minimization, they use (projected) gradient descent to refine the initial solution, and are guaranteed to converge to the global optimum at a linear rate. Notably, the sample complexity of these algorithms does not depend on the target accuracy and is only slightly larger than that of convex programming approaches. Our proposed algorithm also belongs to this line of research but with significant innovations in both algorithm and theory (c.f. Section 1.1).

### 2.2 Matrix Completion with Side Information

Matrix completion with side information has drawn much attention for improving the performance of traditional matrix completion methods in various applications. This method dates back to Jain and Dhillon (2013); Xu et al. (2013), where they proposed the so-called Inductive Matrix Completion methods independently. The method is "inductive", in that it can be generalized to previously unobserved data points, which resolves a major drawback in traditional recommender systems. Extensions to noisy features (Chiang et al., 2015) and non-linear models (Si et al., 2016) have been studied and similar formulation has also been extended to the problem of robust PCA (Chiang et al., 2016; Niranjan et al., 2017; Xue et al., 2017).

Theoretically, side information allows us to recover the target matrix with sample complexity depending on the intrinsic feature dimension rather than the ambient dimension. Information theoretically speaking, with known features, $O(rn)$ samples are sufficient for exact recovery and this is achieved up to some logarithmic factors by the convex relaxation based algorithm proposed in Xu et al. (2013). However, such formulation requires solving a nuclear norm minimization problem and in general cannot have the linear convergence. Jain and Dhillon (2013) adopted ideas from Jain et al. (2013); Hardt (2014); Hardt and Wootters (2014) to obtain a linear convergent algorithm but it requires $O\left(r^3 n^2 \log n \log(1/\epsilon)\right)$ samples. See Table 1 for a detailed comparison between our method and two existing inductive matrix completion algorithms: **Maxide** (Xu et al., 2013) and **AltMin** (Jain and Dhillon, 2013). It is worth noting that our approach achieves both linear rate of convergence and sample complexity linear in the feature dimension $n$.

Table 1: Comparison results of sample complexity and convergence rate for different inductive matrix completion algorithms.

| Algorithm | Sample Complexity | Linear Convergence? |
|:---:|:---:|:---:|
| **Maxide** (Xu et al., 2013) | $O(rn \log n \log d)$ | No |
| **AltMin**[2] (Jain and Dhillon, 2013) | $O(r^3 n^2 \log n \log(1/\epsilon))$ | Yes |
| **Ours** | $O(r^2 n \log n \log d)$ | Yes |

## 3 Problem Setup and Preliminaries

Recall that our goal is to recover the unknown rank-$r$ matrix $\mathbf{L}^* \in \mathbb{R}^{d_1 \times d_2}$ by learning a lower-dimensional matrix $\mathbf{M}^* \in \mathbb{R}^{n_1 \times n_2}$ given the side information in terms of $\mathbf{X}_L$ and $\mathbf{X}_R$. Denote the rank-$r$ singular value decomposition (SVD) of $\mathbf{M}^*$ by $\mathbf{M}^* = \overline{\mathbf{U}}^* \mathbf{\Sigma}^* \overline{\mathbf{V}}^{*\top}$. Let $\sigma_1^* \geq \sigma_2^* \geq \ldots \geq \sigma_r^* > 0$ be the sorted singular values of $\mathbf{M}^*$ and $\kappa = \sigma_1^*/\sigma_r^*$ be the condition number. Assume each entry of $\mathbf{L}^*$ is observed independently with probability $p \in (0,1)$. In particular, for any $(i,j) \in [d_1] \times [d_2]$, we consider the following Bernoulli observation model

$$L_{ij} = \begin{cases} L_{ij}^*, & \text{with probability } p; \\ *, & \text{otherwise.} \end{cases} \tag{3.1}$$

Let $\Omega$ be the index set of observed entries in $\mathbf{L}^*$, i.e., $\Omega = \{(i,j) \in [d_1] \times [d_2] \mid L_{ij} \neq *\}$. Note that restricting on the observed index set $\Omega$, we have $\mathcal{P}_\Omega(\mathbf{L}) = \mathcal{P}_\Omega(\mathbf{L}^*)$.

In order to fully exploit the side information, following Xu et al. (2013); Yi et al. (2013); Chiang et al. (2016), we assume the following standard feasibility condition: $\text{col}(\mathbf{X}_L) \supseteq \text{col}(\mathbf{L}^*)$, $\text{col}(\mathbf{X}_R) \supseteq \text{col}(\mathbf{L}^{*\top})$, where $\text{col}(\mathbf{A})$ represents the column space of matrix $\mathbf{A}$. Intuitively, this condition suggests that the feature matrices are correlated to the underlying true low-rank space, so that we could make use of the feature information to improve our recovery. In other words, we assume $\mathbf{L}^*$ can be decomposed as $\mathbf{L}^* = \mathbf{X}_L \mathbf{M}^* \mathbf{X}_R^\top$. In addition, without loss of generality, we assume both feature matrices $\mathbf{X}_L$ and $\mathbf{X}_R$ have orthonormal columns[3], i.e., $\mathbf{X}_L^\top \mathbf{X}_L = \mathbf{I}_{n_1}, \mathbf{X}_R^\top \mathbf{X}_R = \mathbf{I}_{n_2}$.

---

[2] Jain and Dhillon (2013) requires a weaker incoherence condition in that they only assume the features are incoherent. However, when additional incoherence condition is imposed on the true singular spaces of $\mathbf{L}^*$, it is unclear whether their algorithm can reduce the sample complexity or not.

It is well-known in matrix completion (Gross, 2011) that if $\mathbf{L}^*$ is equal to zero in nearly all of the rows or columns, recovering $\mathbf{L}^*$ exactly is impossible unless all of its entries are sampled. Therefore, we impose the standard incoherence condition on the unknown low-rank matrix $\mathbf{L}^*$ (Candès and Recht, 2009; Recht, 2011; Yi et al., 2016). Note that given feature matrices $\mathbf{X}_L, \mathbf{X}_R$, the singular value decomposition of $\mathbf{L}^*$ can be formulated as $(\mathbf{X}_L \overline{\mathbf{U}}^*) \boldsymbol{\Sigma}^* (\mathbf{X}_R \overline{\mathbf{V}})^\top$.

**Assumption 3.1** (Incoherence for $\mathbf{L}^*$). The unknown low-rank matrix $\mathbf{L}^*$ is $\mu_0$-incoherent, i.e.,

$$\|\mathbf{X}_L \overline{\mathbf{U}}^*\|_{2,\infty} \leq \sqrt{\frac{\mu_0 r}{d_1}} \quad \text{and} \quad \|\mathbf{X}_R \overline{\mathbf{V}}^*\|_{2,\infty} \leq \sqrt{\frac{\mu_0 r}{d_2}}.$$

Furthermore, following Jain and Dhillon (2013); Xu et al. (2013); Chiang et al. (2016), we impose the following incoherence condition on the feature matrices.

**Assumption 3.2** (Incoherence for feature matrices). The feature matrices $\mathbf{X}_L$ and $\mathbf{X}_R$ are both self-inchoherent with parameter $\mu_1$, i.e,

$$\|\mathbf{X}_L\|_{2,\infty} \leq \sqrt{\frac{\mu_1 n_1}{d_1}}, \quad \text{and} \quad \|\mathbf{X}_R\|_{2,\infty} \leq \sqrt{\frac{\mu_1 n_2}{d_2}}.$$

With the aid of additional feature information, inductive matrix completion can be formulated as the following optimization problem

$$\min_{\mathbf{M} \in \mathbb{R}^{n_1 \times n_2}} \frac{1}{2p} \big\| \mathcal{P}_\Omega (\mathbf{X}_L \mathbf{M} \mathbf{X}_R^\top - \mathbf{L}) \big\|_F^2, \quad \text{subject to} \quad \text{rank}(\mathbf{M}) \leq r, \tag{3.2}$$

where $\Omega$ is the index set of observed entries and $p = |\Omega|/(d_1 d_2)$ denotes the sampling probability in the observation model. In order to estimate the low-rank matrix $\mathbf{M}^*$ more efficiently, following Tu et al. (2015), Zheng and Lafferty (2015) and Yi et al. (2016), we propose to solve the following factorized non-convex optimization problem

$$\min_{\substack{\mathbf{U} \in \mathbb{R}^{n_1 \times r} \\ \mathbf{V} \in \mathbb{R}^{n_2 \times r}}} \frac{1}{2p} \big\| \mathcal{P}_\Omega (\mathbf{X}_L \mathbf{U} \mathbf{V}^\top \mathbf{X}_R^\top - \mathbf{L}) \big\|_F^2. \tag{3.3}$$

Due to the reparameterization $\mathbf{M} = \mathbf{U} \mathbf{V}^\top$, the rank constraint in (3.2) is automatically guaranteed in (3.3).

## 4 The Proposed Algorithm

Let $\mathbf{U}^* = \overline{\mathbf{U}}^* \boldsymbol{\Sigma}^{*1/2}$ and $\mathbf{V}^* = \overline{\mathbf{V}}^* \boldsymbol{\Sigma}^{*1/2}$ be the true factorized matrices. It is obvious that $(\mathbf{U}^*, \mathbf{V}^*)$ is the optimal solution to optimization problem (3.3). However, for any invertible matrix $\mathbf{P} \in \mathbb{R}^{r \times r}$, $(\mathbf{U}^* \mathbf{P}, \mathbf{V}^* (\mathbf{P}^{-1})^\top)$ is also an optimal solution. In order to deal with this identifiability issue, following Tu et al. (2015); Zheng and Lafferty (2016); Park et al. (2016), we impose an additional regularizer to the objective function in (3.3) to penalize the scale difference between $\mathbf{U}$ and $\mathbf{V}$. Specifically, we consider the following regularized optimization problem

$$\min_{\substack{\mathbf{U} \in \mathbb{R}^{n_1 \times r} \\ \mathbf{V} \in \mathbb{R}^{n_2 \times r}}} f_\Omega(\mathbf{U}, \mathbf{V}) := \frac{1}{2p} \big\| \mathcal{P}_\Omega (\mathbf{X}_L \mathbf{U} \mathbf{V}^\top \mathbf{X}_R^\top - \mathbf{L}) \big\|_F^2 + \frac{1}{8} \big\| \mathbf{U}^\top \mathbf{U} - \mathbf{V}^\top \mathbf{V} \big\|_F^2, \tag{4.1}$$

where $\mathbf{U} \in \mathbb{R}^{n_1 \times r}$, $\mathbf{V} \in \mathbb{R}^{n_2 \times r}$, and $f_\Omega$ denotes the regularized sample loss function. Intuitively speaking, the regularization term encourages the two factorized matrices $\mathbf{U}$ and $\mathbf{V}$ to have a similar scale.

---

[3] In practice, one could conduct QR factorization or SVD to acquire the corresponding orthonormal feature matrices.

We propose a multi-phase gradient-based algorithm to solve the proposed estimator (4.1), as shown in Algorithm 1. More specifically, we first randomly split the observed index set $\Omega$ into $S+1$ independent subsets $\{\Omega_s\}_{s=0}^{S}$, where $\Omega_0$ has cardinality $|\Omega|/2$ and each of the rest has cardinality $|\Omega|/(2S)$. In Phase 1, we project the observed matrix $\mathbf{L}$ onto the first subset $\Omega_0$ and perform rank-$r$ SVD on $p_0^{-1}\mathcal{P}_{\Omega_0}(\mathbf{L}^*)$ to get an initial estimator $(\mathbf{U}_{\text{init}}, \mathbf{V}_{\text{init}})$, where $p_0 = |\Omega_0|/(d_1 d_2)$. We use $\text{SVD}_r(\cdot)$ to denote the rank-$r$ SVD.

In Phase 2, we perform projected gradient descent with resampling (Jain et al., 2013; Jain and Dhillon, 2013) (a.k.a., sample splitting), where we use one fresh subset for each gradient descent update. The projection step guarantees that each intermediate iterate satisfies the similar incoherence condition as that of $(\mathbf{U}^*, \mathbf{V}^*)$, while the resampling scheme ensures the independence of the samples used in the current iteration and the previous iterates. As will be clear in the next section and in the proofs, the second phase is crucial in reducing the variance of gradient estimate and ensures the uniform convergence in the third phase. The constraint sets $\mathcal{C}_1$ and $\mathcal{C}_2$ associated with the projection are defined as

$$\mathcal{C}_1 = \left\{ \mathbf{U} \in \mathbb{R}^{n_1 \times r} \;\Big|\; \|\mathbf{X}_L \mathbf{U}\|_{2,\infty} \leq \sqrt{\frac{\mu_0 r}{d_1}} \|\mathbf{Z}_{\text{init}}\|_2 \right\}, \quad (4.2)$$
$$\mathcal{C}_2 = \left\{ \mathbf{V} \in \mathbb{R}^{n_2 \times r} \;\Big|\; \|\mathbf{X}_R \mathbf{V}\|_{2,\infty} \leq \sqrt{\frac{\mu_0 r}{d_2}} \|\mathbf{Z}_{\text{init}}\|_2 \right\},$$

where $\mathbf{Z}_{\text{init}}$ is specified in Phase 1. Let $\mathcal{P}_{\mathcal{C}_1}(\widehat{\mathbf{U}})$ be the projection of $\widehat{\mathbf{U}} \in \mathbb{R}^{n_1 \times r}$ onto $\mathcal{C}_1$, which can be alternatively regarded as the exact solution to the following convex quadratically-constrained-quadratic-programming (QCQP)

$$\operatorname*{argmin}_{\mathbf{U} \in \mathbb{R}^{n_1 \times r}} \frac{1}{2}\|\mathbf{U} - \widehat{\mathbf{U}}\|_F^2, \quad \text{subject to } \left\|[\mathbf{X}_L \mathbf{U}]_{i,*}\right\|_2^2 \leq \frac{\mu_0 r}{d_1} \|\mathbf{Z}_{\text{init}}\|_2^2, \forall\, i \in [d_1]. \quad (4.3)$$

It is worth noting that convex QCQP problem can be solved approximately and efficiently using interior point methods (Nemirovskii, 2004). Let $\mathcal{P}_{\mathcal{C}_1}(\widehat{\mathbf{U}}, \delta)$ be the $\delta$-approximate solution to optimization problem (4.3), i.e., $\|\mathcal{P}_{\mathcal{C}_1}(\widehat{\mathbf{U}}, \delta) - \mathcal{P}_{\mathcal{C}_1}(\widehat{\mathbf{U}})\|_F \leq \delta$. Similarly, the QCQP problem with respect to $\mathbf{V}$ is formulated in a similar way, except that $\mathbf{X}_L$ (resp. $d_1$) is replaced with $\mathbf{X}_R$ (resp. $d_2$). Accordingly, we use $\mathcal{P}_{\mathcal{C}_2}(\widehat{\mathbf{V}})$ to denote the exact projection, and $\mathcal{P}_{\mathcal{C}_2}(\widehat{\mathbf{V}}, \delta)$ to be the $\delta$-approximate projection. In addition, the loss function used in the $s$-th iteration of Phase 2 is based on the subset $\Omega_s$, and it is identical to the loss function in (4.1), except that $\Omega$ (resp. $p$) is replaced with $\Omega_s$ (resp. $p_s = |\Omega_s|/(d_1 d_2)$).

Finally, in Phase 3, vanilla gradient descent is performed based on the entire observed matrix $\mathcal{P}_\Omega(\mathbf{L}^*)$. Provided these three phases, as will be seen in later analysis, Algorithm 1 is guaranteed to converge to the true factorized matrices $(\mathbf{U}^*, \mathbf{V}^*)$ with a linear rate of convergence.

## 5 Main Theory

Before presenting the main theoretical results, we note that the optimal solution to optimization problem (4.1) is not unique. Therefore, following Tu et al. (2015), we introduce the so-called Procrustes distance.

**Definition 5.1.** For any $\mathbf{Z} \in \mathbb{R}^{(n_1+n_2) \times r}$, let $D(\mathbf{Z}, \mathbf{Z}^*)$ be the minimal distance between $\mathbf{Z}$ and $\mathbf{Z}^*$ in terms of the optimal rotation, or more precisely, $D(\mathbf{Z}, \mathbf{Z}^*) = \min_{\mathbf{R} \in \mathbb{Q}_r} \|\mathbf{Z} - \mathbf{Z}^* \mathbf{R}\|_F$, where $\mathbb{Q}_r$ denotes the set of $r$-by-$r$ othorgonal matrices and $\mathbf{Z}^* = [\mathbf{U}^*; \mathbf{V}^*]$.

For simplicity, we use $d$ to denote $\max\{d_1, d_2\}$ and $n$ to denote $\max\{n_1, n_2\}$ in the following discussions. In the sequel, we are going to present the theoretical guarantees regarding to the three phases of Algorithms 1.

**Theorem 5.2** (Initialization). Assume the observed index set $\Omega$ follows Bernoulli model (3.1). Suppose Assumptions 3.1, 3.2 hold for the unknown low-rank matrix $\mathbf{L}^*$ and the feature matrices $\mathbf{X}_L, \mathbf{X}_R$, respectively.

**Algorithm 1** GD for IMC
___
**Input:** Observed matrix $\mathcal{P}_\Omega(\mathbf{L}^*)$; feature matrices $\mathbf{X}_L, \mathbf{X}_R$; parameter $p_0 = |\Omega|/(2d_1 d_2)$; step size $\tau, \eta$; number of iterations $S, T$, approximation error $\delta$.

  Randomly split $\Omega$ into subsets $\Omega_0, \Omega_1, \ldots, \Omega_S$ with
  $|\Omega_0| = |\Omega|/2$ and $|\Omega_s| = |\Omega|/(2S)$, for any $s \in [S]$
  **// Phase 1: Initialization**
  $[\widetilde{\mathbf{U}}_0, \boldsymbol{\Sigma}_0, \widetilde{\mathbf{V}}_0] = \text{SVD}_r\left(p_0^{-1}\mathcal{P}_{\Omega_0}(\mathbf{L}^*)\right)$
  $\mathbf{U}_{\text{init}} = \mathbf{X}_L^\top \widetilde{\mathbf{U}}_0 \boldsymbol{\Sigma}_0^{1/2}; \mathbf{V}_{\text{init}} = \mathbf{X}_R^\top \widetilde{\mathbf{V}}_0 \boldsymbol{\Sigma}_0^{1/2}$
  $\mathbf{Z}_{\text{init}} = [\mathbf{U}_{\text{init}}; \mathbf{V}_{\text{init}}]$
  **// Phase 2: PGD with subsamples**
  $\mathbf{U}_0 = \mathcal{P}_{\mathcal{C}_1}(\mathbf{U}_{\text{init}}, \delta), \mathbf{V}_0 = \mathcal{P}_{\mathcal{C}_2}(\mathbf{V}_{\text{init}}, \delta)$
  **for:** $s = 1, 2, \ldots, S$ **do**
    $\mathbf{U}_s = \mathcal{P}_{\mathcal{C}_1}\big(\mathbf{U}_{s-1} - \eta \nabla_\mathbf{U} f_{\Omega_s}(\mathbf{U}_{s-1}, \mathbf{V}_{s-1}), \delta\big)$
    $\mathbf{V}_s = \mathcal{P}_{\mathcal{C}_2}\big(\mathbf{V}_{s-1} - \eta \nabla_\mathbf{V} f_{\Omega_s}(\mathbf{U}_{s-1}, \mathbf{V}_{s-1}), \delta\big)$
  **end for**
  **// Phase 3: Vanilla GD**
  $\mathbf{U}^0 = \mathbf{U}_S, \mathbf{V}^0 = \mathbf{V}_S$
  **for:** $t = 0, 1, \ldots, T-1$ **do**
    $\mathbf{U}^{t+1} = \mathbf{U}^t - \tau \nabla_\mathbf{U} f_\Omega(\mathbf{U}^t, \mathbf{V}^t)$
    $\mathbf{V}^{t+1} = \mathbf{V}^t - \tau \nabla_\mathbf{V} f_\Omega(\mathbf{U}^t, \mathbf{V}^t)$
  **end for**
**Output:** $(\mathbf{U}^T, \mathbf{V}^T)$
___

For any $\gamma \in (0, 1)$, there exist constants $c_1, c_2$ such that under the condition $|\Omega_0| \geq c_1 \mu_0 \mu_1 r^2 \kappa^2 n \log d / \gamma^2$, with probability at least $1 - c_2/d$, the output of Phase 1 in Algorithm 1 satisfies

$$D(\mathbf{Z}_{\text{init}}, \mathbf{Z}^*) \leq 4\gamma \sqrt{\sigma_r^*}.$$

Theorem 5.2 suggests that after the spectral initialization step, the output of Phase 1 is already in a small neighbourhood of the optimum with radius $O(\sqrt{\sigma_r^*})$. Notably, the sample complexity is linear in $n$, in sharp contrast to that of the classical matrix completion setting which is at least linear in $d$.

**Theorem 5.3** (PGD with subsamples). Under the same conditions as in Theorem 5.2, suppose the output of Phase 1, $\mathbf{Z}_{\text{init}}$, satisfies $D(\mathbf{Z}_{\text{init}}, \mathbf{Z}^*) \leq \alpha \sqrt{\sigma_r^*}/2$ with constant $\alpha \leq 1/40$. There exist constants $c_1, c_2, c_3, c_4, c_5$ such that, if the total sample size $|\Omega| \geq c_1 S \cdot \max\{\mu_0 \mu_1 r \kappa n, \mu_0^2 r^2 \kappa^2\} \log d$, with step size $\eta = c_2/(r\sigma_1^*)$ and approximation error $\delta \leq c_3 \sqrt{\sigma_r^*}/(r\kappa)$, the final iterate $(\mathbf{U}_S, \mathbf{V}_S)$ in Phase 2 of Algorithm 1 satisfies

$$D^2(\mathbf{Z}_S, \mathbf{Z}^*) \leq \left(1 - \frac{c_2}{16r\kappa}\right)^S \alpha^2 \sigma_r^* + c_4 \delta r \kappa \sqrt{\sigma_r^*} \tag{5.1}$$

with probability at least $1 - c_5 S/d$, where $\mathbf{Z}_S = [\mathbf{U}_S; \mathbf{V}_S]$.

The last term on the right hand side of (5.1) originates from the approximation error $\delta$ when solving the convex QCQP (4.3) with respect to $\mathbf{U}$ (or $\mathbf{V}$). Theorem 5.3 suggests that under proper initialization, the gradient iteration in Phase 2 converges at a linear rate with contraction parameter $1 - O(1/(r\kappa))$. Note that the step size is chosen as $O(1/(r\sigma_1^*))$. In practice, since $\sigma_1^*$ is unknown, we can approximate $\sigma_1^*$ by $C \cdot \|\mathbf{U}_{\text{init}} \mathbf{V}_{\text{init}}^\top\|_2$ and tune the coefficient $C$.

**Theorem 5.4** (Vanilla GD). Under the same conditions as in Theorem 5.2, suppose the final iterate $(\mathbf{U}_S, \mathbf{V}_S)$ of Phase 2 in Algorithm 1 satisfies $D(\mathbf{Z}_S, \mathbf{Z}^*) \leq c_0 \sqrt{\sigma_r^*}/(\mu_1 n)$ with $\mathbf{Z}_S = [\mathbf{U}_S; \mathbf{V}_S]$ and constant $c_0$ small enough. Then there exist constants $c_1, c_2, c_3$ such that if $|\Omega| \geq c_1 \mu_0 \mu_1 r n \log d$, with step size $\tau = c_2/\sigma_1^*$, the output of Phase 3 in Algorithm 1 satisfies

$$D^2(\mathbf{Z}^T, \mathbf{Z}^*) \leq \left(1 - \frac{\tau \sigma_r^*}{16}\right)^T D^2(\mathbf{Z}_S, \mathbf{Z}^*)$$

with probability at least $1 - c_3/d$, where $\mathbf{Z}^T = [\mathbf{U}^T; \mathbf{V}^T]$.

Theorem 5.4 implies that if the final iterate of Phase 2 falls into a even smaller neighbourhood around the optimum with radius $O(1/n)$, vanilla gradient descent suffices to guarantee the linear rate of convergence.

Putting all the above theorems together, we arrive at the following main result about our algorithm.

**Theorem 5.5.** Assume the observed index set $\Omega$ follows Bernoulli model (3.1) and incoherence Assumptions 3.1, 3.2 hold. There exist constants $c_1, c_2, c_3, c_4$ and $c_5$ such that under condition $|\Omega| \geq c_1 \max\{\mu_1 n, \mu_0 r\kappa\} \mu_0 r^2 \kappa^2 \log n \log d$, if step size $\eta = c_2/(r\sigma_1^*)$, $\tau = c_3/\sigma_1^*$ and approximation error $\delta = O\left(1/(r\kappa n^2)\right)$, after $S = O(r\kappa \log n)$ iterations in Phase 2 and $T = O\left(\kappa \log(1/\epsilon)\right)$ iterations in Phase 3, with probability at least $1 - c_4 r\kappa \log n/d$, the output of Algorithm 1 satisfies

$$\|\mathbf{M}^T - \mathbf{M}^*\|_F \leq c_5 \sqrt{\sigma_1^*} \epsilon,$$

where $\mathbf{M}^T = \mathbf{U}^T \mathbf{V}^{T\top}$ and $\mathbf{M}^* = \mathbf{U}^* \mathbf{V}^{*\top}$.

Theorem 5.5 shows that the overall sample complexity of Algorithm 1 is $O(r^2 \kappa^2 n \log n \log d)$. Here, we explicitly write down the dependency on condition number $\kappa$ in the $O(\cdot)$ notation for completeness. It is worth noting that our gradient-based Algorithm 1 achieves both linear rate of convergence and sample complexity linearly depending on $n$, compared with convex relaxation based approach (Xu et al., 2013) whose convergence rate is sublinear (i.e., $O(1/\sqrt{\epsilon})$) and alternation minimization (Jain and Dhillon, 2013), which requires at least $O(r^3 n^2 \log n \log(1/\epsilon))$ samples.

**Remark 5.6.** The rank-$r$ SVD in Phase 1 requires $O(r|\Omega_0|)$ computation. The runtime of the gradient computation for the $s$-th iteration in the second phase is $O(rn|\Omega_s| + r^2 n)$, while solving the convex QCQP subproblem requires $O(r^2 n^2 d^{3/2} \log d)$ computation if using the path-following interior point method (Nemirovskii, 2004). Thus to perform $S = O(r\kappa \log n)$ iterations, the overall computational complexity of Phase 2 is $O(r^3 n^2 d^{3/2} \log n \log d)$. The runtime of gradient computation in each iteration of Phase 3 is $O(rn|\Omega|)$ and the total number of iterations required in Phase 3 is $T = O(\kappa \log(1/\epsilon))$, which implies the overall computational cost in Phase 3 is $O(r^3 n^2 \log n \log d \log(1/\epsilon))$. Putting all these pieces together, we conclude the total computational complexity of Algorithm 1 is $O(r^3 n^2 \log n \log d \log(1/\epsilon) + r^3 n^2 d^{3/2} \log n \log d)$.

# 6 Experiments

In this section, we compare the proposed gradient-based algorithm with existing inductive matrix completion methods, including the convex relaxation based approach, **Maxide** (Xu et al., 2013) and alternating minimization based algorithm, **AltMin** (Jain and Dhillon, 2013) on both synthetic and real datasets. In addition, the standard matrix completion approach based on non-convex projected gradient descent (Zheng and Lafferty, 2016) (**MC**) is compared as a baseline for simulations and the second real data experiment on gene-disease prediction, while the Binary Relevance approach (Boutell et al., 2004) using linear kernel SVM (Chang and Lin, 2011) (**BR-linear**) is included as a baseline for the first real data experiment on multi-label learning. All algorithms are implemented in Matlab on a machine with Intel 8-core Core i7 3.40 GHz with 8GB RAM.

## 6.1 Simulations

The unknown low-rank matrix $\mathbf{M}^* \in \mathbb{R}^{n_1 \times n_2}$ is generated such that $\mathbf{M}^* = \mathbf{U}^* \mathbf{V}^{*\top}$, and the entries of $\mathbf{U}^* \in \mathbb{R}^{n_1 \times r}, \mathbf{V}^* \in \mathbb{R}^{n_2 \times r}$ are drawn independently from centered Gaussian distribution with variance $1/n_1$ and $1/n_2$, respectively. Let the singular value decomposition of a random matrix $\mathbf{F} \in \mathbb{R}^{d_1 \times d_2}$ be $\mathbf{F} = \mathbf{Y}_L \mathbf{\Sigma} \mathbf{Y}_R^\top$, where each entry of $\mathbf{F}$ is drawn independently from standard normal distribution. The feature matrices $\mathbf{X}_L \in \mathbb{R}^{d_1 \times n_1}, \mathbf{X}_R \in \mathbb{R}^{d_2 \times n_2}$ are then generated as the first $n_1$ columns of $\mathbf{Y}_L$ and the first $n_2$ columns of $\mathbf{Y}_R$, respectively. The observed data matrix $\mathbf{L}$ follows from the Bernoulli model (3.1) with the full data matrix defined by $\mathbf{L}^* = \mathbf{X}_L \mathbf{M}^* \mathbf{X}_R^\top$.

To begin with, we investigate the sample complexity of the proposed gradient-based method under the symmetric setting: $d_1 = d_2 = d$, $n_1 = n_2 = n$. In particular, we consider (i) $d = 500, n = 50, r = 10$; (ii) $d = 500, n = 100, r = 5$; (iii) $d = 1000, n = 50, r = 5$; (iv) $d = 1000, n = 100, r = 10$. We compute the empirical probability of successful recovery after 50 repeated trials, where we regard the trial as successful if the relative error $\|\mathbf{X}_L \mathbf{M}^T \mathbf{X}_R^\top - \mathbf{L}^*\|_F / \|\mathbf{L}^*\|_F$ is less than $10^{-6}$. The experimental results are shown in Figure 1(a). Here, $m$ represents the total number of observed entries. Under all of the aforementioned settings, the phase transition happens to be around $m/(nr) = 6$, which implies that the optimal sample complexity for gradient-based inductive matrix completion approach may be linear in both $n$ and $r$.

Moreover, we compare our algorithm with the aforementioned algorithms, including **MC**, **Maxide** and **AltMin**. All the parameters, such as step size and regularization parameters, are tuned by 5-fold cross validation. We measure the performance by the relative reconstruction error $\|\widehat{\mathbf{L}} - \mathbf{L}^*\|_F / \|\mathbf{L}^*\|_F$ under (i) symmetric setting: $d_1 = d_2 = d = 1000$, $n_1 = n_2 = n = 100$, $r = 10$ with sampling rate $p$ varied in the range $\{2\%, 5\%, 10\%\}$; (ii) rectangular setting: $d_1 = 5000, d_2 = 2000, n_1 = 100, n_2 = 50$, $r = 5$ with $p \in \{0.25\%, 0.5\%, 1\%, 2\%\}$. The simulation results are displayed in Figures 1(b)-1(h). Here, each effective data pass evaluates $|\Omega|$ observed entries. It can be seen that inductive methods can recover the unknown low-rank matrix $\mathbf{L}^*$ successfully using less observed entries compared with the standard matrix completion approach, which proves the effectiveness of feature information. In addition, our approach achieves the lowest recovery error with respect to the same number of effective data passes, and outperforms existing inductive matrix completion algorithms by a large margin. We also demonstrate the relative error versus CPU time plots under the previously mentioned settings based on different (inductive) matrix completion methods. The experimental results are presented in Figure 2, where similar trend as Figure 1 can be observed. All these comparison results clearly demonstrate the superiority of our proposed algorithm in terms of computation and is well aligned with our theory.

## 6.2 Multi-Label Learning

We also apply our proposed algorithm to multi-label learning on the image classification dataset **NUS-WIDE-OBJECT** obtained from Chua et al. (2009), which is one of the prominent applications of inductive matrix completion. The NUS-WIDE-OBJECT dataset consists of $d_1 = 30000$ images classified by $d_2 = 31$ object categories, along with 5 types of low-level features extracted from these images. We construct the feature matrix by further extracting the top-50 principle components from each type of side information, which leads to $n_1 = 250$ features in total. Detailed information regarding the dataset can be found in Chua et al. (2009). Our goal is to predict the labels associated with the unseen instances, based on both the side information as well as the label assignments of the observed instances. By leveraging the low-rankness property of the unknown instance-label matrix (Ji et al., 2008; Goldberg et al., 2010), multi-label learning can be reformulated as an inductive matrix completion problem (3.2), where $\mathbf{L}^*$ corresponds to the instance-label matrix, $\mathbf{X}_L$ represents the feature matrix and $\mathbf{X}_R$ is set as an identity matrix in this context.

We randomly sample $p \times 100\%$ instances as the observed (training) data for each dataset, and treat the remaining $(1-p) \times 100\%$ instances as the unobserved (testing) data, with $p$ chosen from $\{10\%, 25\%, 50\%\}$. We estimate the unknown matrix of parameters based on the training data, and report the average precision (AP) (Zhang and Zhou, 2014) computed from the testing data. Specifically, the average precision measures the averaged fraction of relevant labels ranked higher than a specific label. We compare our algorithm with the baseline approach, **BR-linear**, and existing inductive matrix completion algorithms, **Maxide** and **AltMin**. All the parameters, including the rank $r$ (we tune it over the grid $\{5, 10, \ldots, 30\}$), are tuned via 5-fold cross validation based on the training data. Table 2 depicts the detailed experimental results. In detail, for each setting of observed training data, we report the averaged AP over 10 trials and the corresponding standard deviation as well as the total run time. We can observe from Table 2 that the proposed gradient-based algorithm outperforms the BR-linear by a large margin. Compared with existing inductive matrix completion algorithms, our algorithm also achieves significantly better results under all of the experimental settings in terms of both prediction accuracy and running time. This again illustrates the advantage of our algorithm.

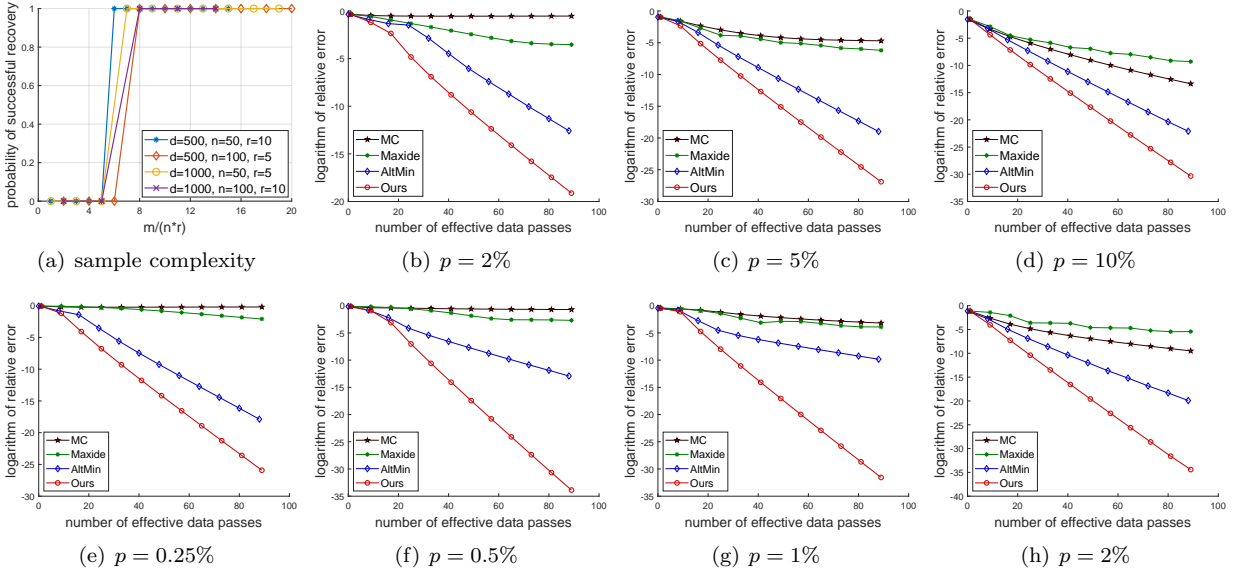

Figure 1: Experimental results on synthetic datasets: (a) Plot of empirical probability of successful recovery versus $m/(nr)$ based on our proposed algorithm under the setting $d = 1000$, $n = 100$ and $r = 10$. (b),(c),(d) Plots of logarithm relative error versus effective data passes for different (inductive) matrix completion algorithms under the symmetric setting $d = 1000$, $n = 100$ and $r = 10$ with different sampling rate $p$. (e),(f),(g),(h) Plots of logarithm relative error vs. effective data passes for different (inductive) matrix completion methods under the rectangular setting that $d_1 = 5000$, $d_2 = 2000$, $n_1 = 100$, $n_2 = 50$ and $r = 5$ with varied sampling rate.

Table 2: Experimental results in terms of AP and total running time on NUS-WIDE-OBJECT dataset for multi-label learning via different methods. $p$ represents the percentage of observed instances. The best averaged AP (the higher the better) is bolded for each setting.

| Method | $p = 10\%$ | | $p = 25\%$ | | $p = 50\%$ | |
|---|---|---|---|---|---|---|
| | averaged AP (std) | time (s) | averaged AP (std) | time (s) | averaged AP (std) | time (s) |
| BR-linear | 0.3280 (0.0037) | $9.72 \times 10^2$ | 0.3357 (0.0046) | $2.77 \times 10^3$ | 0.3428 (0.0031) | $7.15 \times 10^3$ |
| Maxide | 0.5349 (0.0034) | $3.21 \times 10^1$ | 0.5562 (0.0021) | $3.42 \times 10^1$ | 0.5629 (0.0023) | $3.27 \times 10^1$ |
| AltMin | 0.5265 (0.0031) | $1.92 \times 10^1$ | 0.5536 (0.0028) | $2.11 \times 10^1$ | 0.5591 (0.0027) | $2.03 \times 10^1$ |
| Ours | **0.5434** (0.0040) | $7.53 \times 10^0$ | **0.5677** (0.0027) | $7.19 \times 10^0$ | **0.5718** (0.0023) | $9.57 \times 10^0$ |

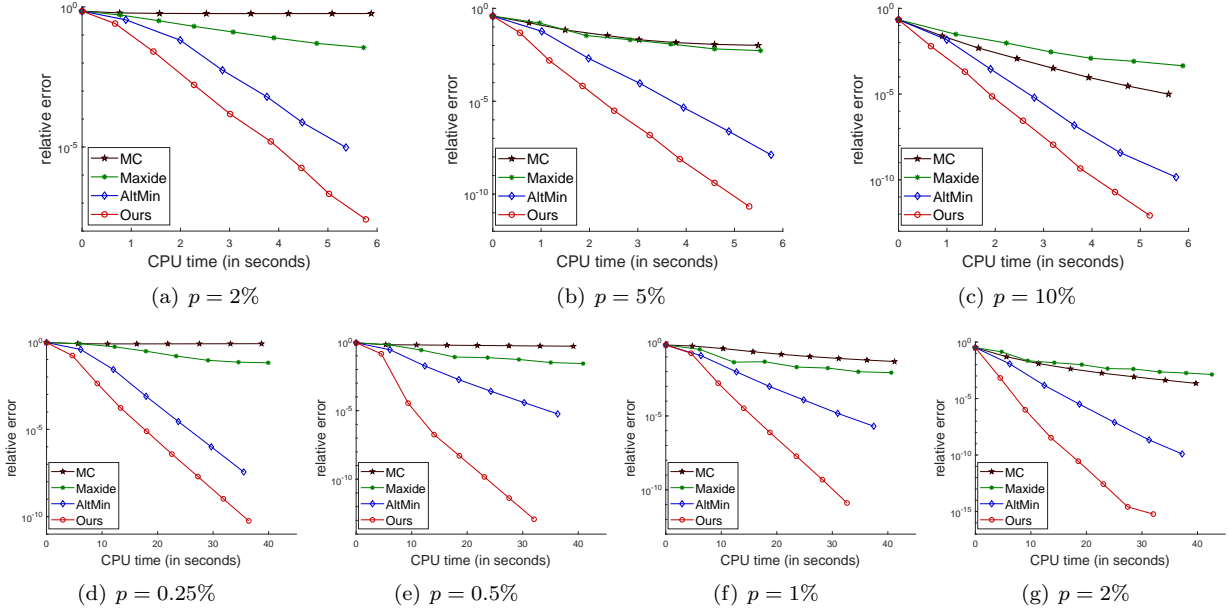

Figure 2: Plots of relative error vs. CPU time for different (inductive) matrix completion algorithms under the settings: (i) $d = 1000$, $n = 100$ and $r = 10$ with sampling rate $p$ selected from $\{2\%, 5\%, 10\%\}$ in the top panel. (ii) $d_1 = 5000$, $d_2 = 2000$, $n_1 = 100$, $n_2 = 50$ and $r = 5$ with $p$ varied in the range $\{0.25\%, 0.5\%, 1\%, 2\%\}$ in the bottom panel.

### 6.3 Gene-Disease Prediction

We further apply our proposed method for predicting gene-disease associations on the OMIM[4] data used in Singh-Blom et al. (2013), which is another successful application of inductive matrix completion. In the context of gene-disease association prediction, we let $\mathbf{L}^* \in \mathbb{R}^{d_1 \times d_2}$ be the gene-disease association matrix, such that $L^*_{ij} = 1$ if gene $i$ is associated with disease $j$; $L^*_{ij} = 0$ if the association is unobserved. On this dataset, the association matrix $\mathbf{L}^*$ is highly sparse, consisting of $d_1 = 12331$ different genes and $d_2 = 3209$ different diseases with only 3954 discovered gene-disease associations. In addition, we obtain the gene feature matrix $\mathbf{X}_L \in \mathbb{R}^{d_1 \times n_1}$ and disease feature matrix $\mathbf{X}_R \in \mathbb{R}^{d_2 \times n_2}$ from Natarajan and Dhillon (2014), where $n_1 = 300$ gene features and $n_2 = 200$ disease features are extracted respectively. Our objective is to predict potential genes for certain diseases of interest based on both the observed associations and feature information, which can thus be formulated as an inductive matrix completion problem. Following Natarajan and Dhillon (2014), we solve the following optimization problem (4.1) with an additional regularizer to take into account the sparsity of the underlying association matrix

$$\min_{\substack{\mathbf{U} \in \mathbb{R}^{n_1 \times r} \\ \mathbf{V} \in \mathbb{R}^{n_2 \times r}}} f_\Omega(\mathbf{U}, \mathbf{V}) + \lambda \, \|\mathcal{P}_{\Omega^c}(\mathbf{X}_L \mathbf{U} \mathbf{V}^\top \mathbf{X}_R^\top)\|_F^2, \tag{6.1}$$

where $r$ is the supposed rank of $\mathbf{L}^*$, $\Omega$ stands for the (training) index set of gene-disease associations, and $\Omega^c$ represents its complement. Note that all the algorithms we studied here including ours can be directly applied to solve (6.1) with slight modification. In our experiment, we tune the regularization parameter $\lambda$ via cross validation and choose the best value $\lambda = 0.5$.

To evaluate the performance of our method, we equally split the known gene-disease associations into three groups and perform 3-fold cross validation. Specifically, we treat each group as testing data once and apply our gradient-based method on the remaining two groups to obtain the estimation matrix of $\mathbf{L}^*$. For every gene-disease pair $(g, d)$ in the testing group, we order all the genes by the corresponding estimated values associated with disease $d$, and then record the ranking of gene $g$ in the list. We use the cumulative distribution of the rankings (Singh-Blom et al., 2013; Natarajan and Dhillon, 2014) as the performance measure for evaluation, i.e., the probability that the ranking is less than a specific threshold $k \in \{1, 2, \ldots, 100\}$. The experimental results with rank $r$ varied in the range $\{10, 30, 50, 100, 200\}$ based on our method are displayed in Figure 3(a), which indicates that the rank plays an important role in gene-disease prediction: higher rank leads to better performance. In later experiments, we choose $r = 200$ because the performance of inductive matrix completion on this dataset tends to be saturated when $r = 200$.

---
[4]OMIM is short for Online Mendelian Inheritance in Man, which is a public database for human gene-disease studies.

Moreover, we compare our algorithm with the following algorithms: **MC**, **Maxide** and **AltMin**, which are discussed at the beginning of Section 6. The comparison results in terms of the cumulative distribution of the rankings are illustrated in Figure 3(b). It can be seen that our proposed algorithm uniformly outperforms other methods over all threshold values $k$. In addition, we present the precision-recall curves for all the methods we compared in Figure 3(c). Here the precision is defined as the ratio of true recovered gene-disease associations to the total number of associations we assessed; and the recall is the fraction of the true gene-disease associations that are recovered. Again, the proposed method dominates other relevant approaches, which suggests that our method can better serve for biologists to discover new gene-disease associations.

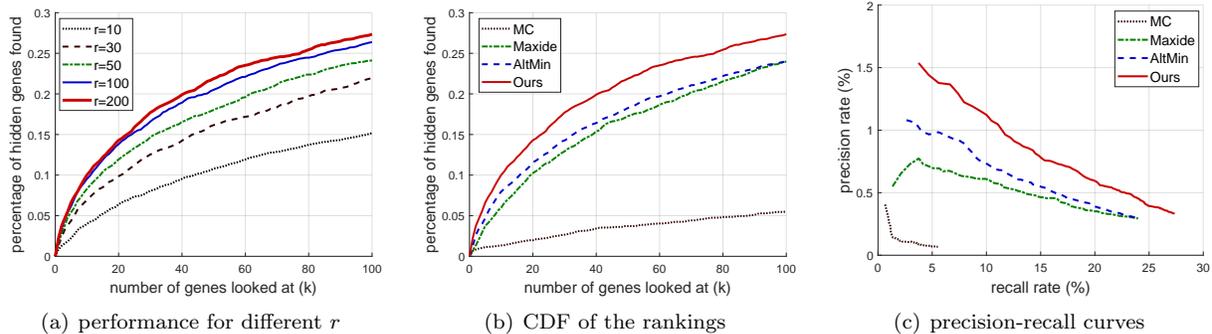

(a) performance for different $r$  (b) CDF of the rankings  (c) precision-recall curves

Figure 3: Experimental results for predicting gene-disease associations: (a) Plot of the probability that a true gene-disease association is recovered in the top-$k$ predictions based our proposed method for different rank $r$; (b) Comparisons of different (inductive) matrix completion methods based on the empirical cumulative distribution of the rankings with rank $r = 200$; (c) Comparisons of different (inductive) matrix completion methods with respect to the standard precision recall measures with rank $r = 200$.

## 7 Conclusions and Future Work

In this paper we proposed the first gradient-based non-convex optimization algorithm for inductive matrix completion with sample complexity linear in the number of features and converges to the unknown low-rank matrix at a linear rate. One possible future direction is to extend our algorithm to the case with noisy side information (Chiang et al., 2015) or the agnostic setting, i.e., the underlying matrix is high-rank (Du et al., 2017b). Another interesting direction is to generalize our approach to non-linear models (Si et al., 2016).

## 8 Acknowledgment

The authors would like to thank Yining Wang for insightful discussions.

# A  Proofs of the Main Results

In this section, we provide the proofs for our main theoretical results. To begin with, we introduce some notations to simplify our proof. Let $\mathcal{I}: \mathbb{R}^{d_1 \times d_2} \to \mathbb{R}^{d_1 \times d_2}$ be the identity map, i.e., $\mathcal{I}(\mathbf{A}) = \mathbf{A}$. Denote the elementwise infinity norm of matrix $\mathbf{A}$ by $\|\mathbf{A}\|_{\infty,\infty}$. For any $\mathbf{Z} \in \mathbb{R}^{(n_1+n_2) \times r}$, we denote $\mathbf{Z} = [\mathbf{U}; \mathbf{V}]$, where $\mathbf{U} \in \mathbb{R}^{n_1 \times r}$ and $\mathbf{V} \in \mathbb{R}^{n_2 \times r}$. According to (4.1), our objective is equivalent to minimize the following regularized loss function in terms of $\mathbf{Z}$

$$\widetilde{f}_\Omega(\mathbf{Z}) := f_\Omega(\mathbf{U}, \mathbf{V}) = \frac{1}{2p}\|\mathcal{P}_\Omega(\mathbf{X}_L \mathbf{U} \mathbf{V}^\top \mathbf{X}_R^\top - \mathbf{L})\|_F^2 + \frac{1}{8}\|\mathbf{U}^\top \mathbf{U} - \mathbf{V}^\top \mathbf{V}\|_F^2. \tag{A.1}$$

Let $\mathrm{Sym}: \mathbb{R}^{d_1 \times d_2} \to \mathbb{R}^{(d_1+d_2) \times (d_1+d_2)}$ be the lifting operator, such that for any matrix $\mathbf{A} \in \mathbb{R}^{d_1 \times d_2}$

$$\mathrm{Sym}(\mathbf{A}) = \begin{bmatrix} \mathbf{0} & \mathbf{A} \\ \mathbf{A}^\top & \mathbf{0} \end{bmatrix}.$$

For any block matrices $\mathbf{A} \in \mathbb{R}^{(d_1+d_2) \times (d_1+d_2)}$ with partitions

$$\mathbf{A} = \begin{bmatrix} \mathbf{A}_{11} & \mathbf{A}_{12} \\ \mathbf{A}_{21} & \mathbf{A}_{22} \end{bmatrix}, \text{ where } \mathbf{A}_{ij} \in \mathbb{R}^{d_i \times d_j}, i,j \in \{1,2\},$$

define linear operators $\mathcal{P}_{\mathrm{diag}}$ and $\mathcal{P}_{\mathrm{off}}: \mathbb{R}^{(d_1+d_2) \times (d_1+d_2)} \to \mathbb{R}^{(d_1+d_2) \times (d_1+d_2)}$ as

$$\mathcal{P}_{\mathrm{diag}}(\mathbf{A}) = \begin{bmatrix} \mathbf{A}_{11} & \mathbf{0} \\ \mathbf{0} & \mathbf{A}_{22} \end{bmatrix} \quad \text{and} \quad \mathcal{P}_{\mathrm{off}}(\mathbf{A}) = \begin{bmatrix} \mathbf{0} & \mathbf{A}_{12} \\ \mathbf{A}_{21} & \mathbf{0} \end{bmatrix}.$$

Similarly, for any block matrices $\mathbf{B} \in \mathbb{R}^{(n_1+n_2) \times (n_1+n_2)}$ with partitions

$$\mathbf{B} = \begin{bmatrix} \mathbf{B}_{11} & \mathbf{B}_{12} \\ \mathbf{B}_{21} & \mathbf{B}_{22} \end{bmatrix}, \text{ where } \mathbf{B}_{ij} \in \mathbb{R}^{n_i \times n_j}, i,j \in \{1,2\},$$

define operators $\overline{\mathcal{P}}_{\mathrm{diag}}$ and $\overline{\mathcal{P}}_{\mathrm{off}}: \mathbb{R}^{(n_1+n_2) \times (n_1+n_2)} \to \mathbb{R}^{(n_1+n_2) \times (n_1+n_2)}$ as

$$\overline{\mathcal{P}}_{\mathrm{diag}}(\mathbf{B}) = \begin{bmatrix} \mathbf{B}_{11} & \mathbf{0} \\ \mathbf{0} & \mathbf{B}_{22} \end{bmatrix} \quad \text{and} \quad \overline{\mathcal{P}}_{\mathrm{off}}(\mathbf{B}) = \begin{bmatrix} \mathbf{0} & \mathbf{B}_{12} \\ \mathbf{B}_{21} & \mathbf{0} \end{bmatrix}.$$

In addition, let $\widetilde{\Omega} \in [d_1 + d_2] \times [d_1 + d_2]$ be the corresponding index set of observed entries in the lifted space, then the observed matrix in the lifted space can be expressed as

$$\mathcal{P}_{\widetilde{\Omega}}(\mathrm{Sym}(\mathbf{L})) = \begin{bmatrix} \mathbf{0} & \mathcal{P}_\Omega(\mathbf{L}) \\ (\mathcal{P}_\Omega(\mathbf{L}))^\top & \mathbf{0} \end{bmatrix}.$$

And we let $\mathbf{X} \in \mathbb{R}^{(d_1+d_2) \times (n_1+n_2)}$ be the corresponding feature matrix in the lifted space, such that

$$\mathbf{X} = \begin{bmatrix} \mathbf{X}_L & \mathbf{0} \\ \mathbf{0} & \mathbf{X}_R \end{bmatrix}.$$

Thus with the notations above, the regularized loss function $\widetilde{f}_\Omega$ in (A.1) can be rewritten as

$$\widetilde{f}_\Omega(\mathbf{Z}) = \frac{1}{4p}\|\mathcal{P}_{\widetilde{\Omega}}(\mathbf{X}\mathbf{Z}\mathbf{Z}^\top \mathbf{X}^\top - \mathrm{Sym}(\mathbf{L}))\|_F^2 + \frac{1}{8}\|\mathbf{Z}^\top \mathbf{D} \mathbf{Z}\|_F^2, \tag{A.2}$$

where $\mathbf{D}$ is defined as

$$\mathbf{D} = \begin{bmatrix} \mathbf{I}_{n_1} & \mathbf{0} \\ \mathbf{0} & -\mathbf{I}_{n_2} \end{bmatrix}.$$

Recall that $\mathbf{Z}^* = [\mathbf{U}^*; \mathbf{V}^*]$ and $\mathcal{P}_\Omega(\mathbf{L}) = \mathcal{P}_\Omega(\mathbf{L}^*)$, then the gradient of $\widetilde{f}_\Omega$ can be formulated as

$$\nabla \widetilde{f}_\Omega(\mathbf{Z}) = \frac{1}{p}\mathbf{X}^\top \mathcal{P}_{\widetilde{\Omega}}(\mathbf{X}[\mathbf{Z}\mathbf{Z}^\top - \mathbf{Z}^*\mathbf{Z}^{*\top}]\mathbf{X}^\top)\mathbf{X}\mathbf{Z}$$
$$+ \frac{1}{2}(\overline{\mathcal{P}}_{\mathrm{diag}} - \overline{\mathcal{P}}_{\mathrm{off}})(\mathbf{Z}\mathbf{Z}^T)\mathbf{Z}. \tag{A.3}$$

## A.1 Proof of Theorem 5.2

*Proof.* According to the initialization phase of Algorithm 1, we have

$$\|\mathbf{U}_{\text{init}}\mathbf{V}_{\text{init}}^\top - \mathbf{M}^*\|_2 = \|p_0^{-1}\mathbf{X}_L^\top \mathcal{P}_{\Omega_0}(\mathbf{L}^*)\mathbf{X}_R - \mathbf{M}^*\|_2$$
$$= \|\mathbf{X}_L^\top (p_0^{-1}\mathcal{P}_{\Omega_0}(\mathbf{L}^*) - \mathbf{L}^*)\mathbf{X}_R\|_2$$
$$= \left\|\sum_{(i,j)=(1,1)}^{(d_1,d_2)} (p_0^{-1}\xi_{ij} - 1)L_{ij}^* \cdot \mathbf{X}_L^\top \mathbf{e}_i \mathbf{e}_j^\top \mathbf{X}_R\right\|_2 := \left\|\sum_{(i,j)=(1,1)}^{(d_1,d_2)} \mathbf{S}_{ij}\right\|_2, \quad (A.4)$$

where $\xi_{ij} = 1$, if $(i,j) \in \Omega_0$; $\xi_{ij} = 0$, otherwise. Next we are going to apply Matrix Bernstein to the right hand side of (A.4). Note that $\mathbb{E}[\mathbf{S}_{ij}] = \mathbf{0}$, and we have

$$\|\mathbf{S}_{ij}\|_2 \leq \frac{1}{p_0} |L_{ij}^*| \cdot \|\mathbf{X}_L^\top \mathbf{e}_i\|_2 \cdot \|\mathbf{X}_R^\top \mathbf{e}_j\|_2 \leq \frac{1}{p_0} \|\mathbf{L}^*\|_{\infty,\infty} \cdot \|\mathbf{X}_L\|_{2,\infty} \cdot \|\mathbf{X}_R\|_{2,\infty}.$$

By Assumptions 3.1 and 3.2, we further obtain

$$\|\mathbf{S}_{ij}\|_2 \leq \frac{1}{p_0} \|\mathbf{X}_L\overline{\mathbf{U}}^*\|_{2,\infty} \cdot \|\mathbf{\Sigma}^*\|_2 \cdot \|\mathbf{X}_R\overline{\mathbf{V}}^*\|_{2,\infty} \cdot \sqrt{\frac{\mu_1^2 n_1 n_2}{d_1 d_2}} \leq \frac{\mu_0 \mu_1 r \sqrt{n_1 n_2}\sigma_1^*}{p_0 d_1 d_2}.$$

Denote $\mathbf{S} = \sum_{i,j} \mathbf{S}_{ij}$. To apply Matrix Bernstein, it remains to bound the terms $\|\mathbb{E}(\mathbf{S}\mathbf{S}^\top)\|_2$ and $\|\mathbb{E}(\mathbf{S}^\top \mathbf{S})\|_2$ respectively. Since $\mathbf{S}_{ij}$'s are independent, we have

$$\|\mathbb{E}(\mathbf{S}\mathbf{S}^\top)\|_2 = \left\|\sum_{(i,j)=(1,1)}^{(d_1,d_2)} \mathbb{E}(\mathbf{S}_{ij}\mathbf{S}_{ij}^\top)\right\|_2$$
$$= \frac{1-p_0}{p_0} \left\|\sum_{(i,j)=(1,1)}^{(d_1,d_2)} L_{ij}^{*2} (\mathbf{X}_L^\top \mathbf{e}_i) \cdot \|\mathbf{X}_R^\top \mathbf{e}_j\|_2^2 \cdot (\mathbf{e}_i^\top \mathbf{X}_L)\right\|_2$$
$$\leq \frac{1}{p_0} \left\|\sum_{i=1}^{d_1} \mathbf{e}_i \mathbf{e}_i^\top \cdot \Big[\sum_{j=1}^{d_2} L_{ij}^{*2} \cdot \|\mathbf{X}_R^\top \mathbf{e}_j\|_2^2\Big]\right\|_2$$
$$\leq \frac{1}{p_0} \cdot \max_{i \in [d_1]} \left(\sum_{j=1}^{d_2} L_{ij}^{*2} \cdot \|\mathbf{X}_R^\top \mathbf{e}_j\|_2^2\right) \leq \frac{\mu_1 n_1}{p_0 d_2} \cdot \|\mathbf{L}^*\|_{2,\infty}^2, \quad (A.5)$$

where the first inequality is due to the fact that $\|\mathbf{AB}\|_2 \leq \|\mathbf{A}\|_2 \cdot \|\mathbf{B}\|_2$ and $\mathbf{X}_L$ is orthonormal, the last inequality follows from Assumption 3.2. According to the SVD of $\mathbf{L}^*$ and Assumption 3.1, we have

$$\|\mathbf{L}^*\|_{2,\infty} = \|(\mathbf{X}_L\overline{\mathbf{U}}^*) \cdot \mathbf{\Sigma}^*\|_{2,\infty} \leq \|\mathbf{X}_L\overline{\mathbf{U}}^*\|_{2,\infty} \cdot \|\mathbf{\Sigma}^*\|_2 \leq \sqrt{\frac{\mu_0 r}{d_1}}\sigma_1^*. \quad (A.6)$$

Therefore, plugging (A.6) into (A.5), we obtain

$$\|\mathbb{E}(\mathbf{S}\mathbf{S}^\top)\|_2 \leq \frac{\mu_0 \mu_1 r n_1}{p_0 d_1 d_2} \sigma_1^{*2}.$$

Similarly, we can obtain the upper bound of $\|\mathbb{E}(\mathbf{S}^\top \mathbf{S})\|_2$, which implies

$$\max\left\{\|\mathbb{E}(\mathbf{S}^\top \mathbf{S})\|_2, \|\mathbb{E}(\mathbf{S}\mathbf{S}^\top)\|_2\right\} \leq \frac{\mu_0 \mu_1 r (n_1 + n_2)}{p_0 d_1 d_2} \sigma_1^{*2}.$$

Applying Matrix Bernstein Lemma D.1, under condition that $p_0 \geq c\mu_0\mu_1 r^2\kappa^2(n_1+n_2)\log d/(\gamma^2 d_1 d_2)$, we have

$$\mathbb{P}\left\{\left\|\sum_{(i,j)=(1,1)}^{(d_1,d_2)} \mathbf{S}_{ij}\right\|_2 \geq \gamma \cdot \frac{\sigma_r^*}{\sqrt{r}}\right\} \leq (d_1+d_2)\cdot \exp(-c'\log d) \leq \frac{1}{d},$$

where $c,c',\gamma > 0$ are some constants. This further implies that with probability at least $1 - 1/d$, we have $\|\mathbf{U}_{\text{init}}\mathbf{V}_{\text{init}}^\top - \mathbf{M}^*\|_2 \leq \gamma\sigma_r^*/\sqrt{r}$. Finally, according to Lemma 5.14 in Tu et al. (2015), we obtain

$$D^2(\mathbf{Z}_{\text{init}}, \mathbf{Z}^*) \leq \frac{2}{\sqrt{2}-1}\cdot\frac{\|\mathbf{U}_{\text{init}}\mathbf{V}_{\text{init}}^\top - \mathbf{M}^*\|_F^2}{\sigma_r(\mathbf{M}^*)} \leq \frac{10r\|\mathbf{U}_{\text{init}}\mathbf{V}_{\text{init}}^\top - \mathbf{M}^*\|_2^2}{\sigma_r^*} \leq 10\gamma^2\sigma_r^*,$$

where the second inequality holds because $\text{rank}(\mathbf{U}_{\text{init}}\mathbf{V}_{\text{init}}^\top - \mathbf{M}^*)$ is at most $2r$. □

## A.2 Proof of Theorem 5.3

Before proceeding to the main proof, we introduce the following notations and facts. Recall that $\mathbf{M}^* = \overline{\mathbf{U}}^*\mathbf{\Sigma}^*\overline{\mathbf{V}}^{*\top}$ and $\mathbf{Z}^* = [\overline{\mathbf{U}}^*; \overline{\mathbf{V}}^*]\mathbf{\Sigma}^{*1/2}$, we denote $\widetilde{\mathbf{Z}}^* = [\overline{\mathbf{U}}^*; -\overline{\mathbf{V}}^*]\mathbf{\Sigma}^{*1/2}$. Note that $\mathbf{M}^*$ and $\mathbf{L}^*$ have the same set of singular values, thus for any $\ell \in [r]$, $\sigma_\ell^2(\mathbf{Z}^*) = \sigma_\ell^2(\widetilde{\mathbf{Z}}^*) = 2\sigma_\ell^*$. We further note that $\mathbf{Z}^{*\top}\widetilde{\mathbf{Z}}^* = \widetilde{\mathbf{Z}}^{*\top}\mathbf{Z}^* = \mathbf{0}$, and $\text{Sym}(\mathbf{M}^*) = (\mathbf{Z}^*\mathbf{Z}^{*\top} - \widetilde{\mathbf{Z}}^*\widetilde{\mathbf{Z}}^{*\top})/2$. Define reference function $G(\mathbf{Z})$ as $G(\mathbf{Z}) = \|\mathbf{Z}\mathbf{Z}^\top - \mathbf{Z}^*\mathbf{Z}^{*\top}\|_F^2/4$, then the gradient of G is given by

$$\nabla G(\mathbf{Z}) = (\mathbf{Z}\mathbf{Z}^\top - \mathbf{Z}^*\mathbf{Z}^{*\top})\mathbf{Z}. \tag{A.7}$$

Thus according to (A.3), we have

$$\nabla \widetilde{f}_\Omega(\mathbf{Z}) = \frac{1}{2}\nabla G(\mathbf{Z}) + \frac{1}{2}(\widetilde{\mathbf{Z}}^*\widetilde{\mathbf{Z}}^{*\top})\mathbf{Z} + \mathbf{X}^\top\left(\frac{1}{p}\mathcal{P}_{\widetilde{\Omega}} - \mathcal{P}_{\text{off}}\right)(\mathbf{X}[\mathbf{Z}\mathbf{Z}^\top - \mathbf{Z}^*\mathbf{Z}^{*\top}]\mathbf{X}^\top)\mathbf{X}\mathbf{Z}. \tag{A.8}$$

The following lemmas demonstrate the local curvature and local smoothness properties of $\widetilde{f}_\Omega$, which are proved in Sections B.1 and B.2, respectively. In both lemmas, for any $\mathbf{Z} \in \mathbb{R}^{(n_1+n_2)\times r}$, we let $\mathbf{R} = \arg\min_{\widehat{\mathbf{R}}\in\mathbb{Q}_r}\|\mathbf{Z} - \mathbf{Z}^*\widehat{\mathbf{R}}\|_F$, and denote $\mathbf{H} = \mathbf{Z} - \mathbf{Z}^*\mathbf{R}$.

**Lemma A.1** (Local curvature). Under the previously stated assumptions in Theorem 5.3, for any fixed $\mathbf{Z} = [\mathbf{U};\mathbf{V}] \in \mathbb{R}^{(n_1+n_2)\times r}$ such that $D(\mathbf{Z},\mathbf{Z}^*) \leq \sqrt{2\sigma_r^*/5}$ and $\|\mathbf{X}_L\mathbf{U}\|_{2,\infty} \leq 2\sqrt{\mu_0 r\sigma_1^*/d_1}$, $\|\mathbf{X}_R\mathbf{V}\|_{2,\infty} \leq 2\sqrt{\mu_0 r\sigma_1^*/d_2}$, there exists constants $c_1, c_2$ such that if $|\Omega| \geq c_1\max\{\mu_0^2 r^2\kappa^2, \mu_0\mu_1 r\kappa n\}\log d$, then with probability at least $1 - c_2/d$, we have

$$\langle \nabla\widetilde{f}_\Omega(\mathbf{Z}), \mathbf{H}\rangle \geq \frac{1}{20}\|\mathbf{Z}\mathbf{Z}^\top - \mathbf{Z}^*\mathbf{Z}^{*\top}\|_F^2 + \frac{1}{4\sigma_1^*}\|\widetilde{\mathbf{Z}}^*\widetilde{\mathbf{Z}}^{*\top}\mathbf{Z}\|_F^2 + \frac{\sigma_r^*}{8}\|\mathbf{H}\|_F^2 - 40\|\mathbf{H}\|_F^4,$$

where $\widetilde{\mathbf{Z}}^* = [\mathbf{U}^*; -\mathbf{V}^*]$.

**Lemma A.2** (Local smoothness). Under the previous stated assumptions as in Theorem 5.3, for any fixed $\mathbf{Z} = [\mathbf{U};\mathbf{V}] \in \mathbb{R}^{(n_1+n_2)\times r}$ such that $D(\mathbf{Z},\mathbf{Z}^*) \leq \sqrt{\sigma_r^*}/4$ and $\|\mathbf{X}_L\mathbf{U}\|_{2,\infty} \leq 2\sqrt{\mu_0 r\sigma_1^*/d_1}$, $\|\mathbf{X}_R\mathbf{V}\|_{2,\infty} \leq 2\sqrt{\mu_0 r\sigma_1^*/d_2}$, there exist constants $c_1, c_2$ such that if $|\Omega| \geq c_1\mu_0\mu_1 r\kappa n\log d$, then with probability at least $1 - c_2/d$, we have

$$\|\nabla\widetilde{f}_\Omega(\mathbf{Z})\|_F^2 \leq (16r+4)\sigma_1^*\|\mathbf{Z}\mathbf{Z}^\top - \mathbf{Z}^*\mathbf{Z}^{*\top}\|_F^2 + \|\widetilde{\mathbf{Z}}^*\widetilde{\mathbf{Z}}^{*\top}\mathbf{Z}\|_F^2 + 4r\sigma_r^{*2}\|\mathbf{H}\|_F^2,$$

where $\widetilde{\mathbf{Z}}^* = [\mathbf{U}^*; -\mathbf{V}^*]$.

Now we are ready to prove Theorem 5.3.

*Proof.* Theorem 5.3 will be proved by induction. Consider the $s$-th iteration in Phase 2 of Algorithm 1, for any $s \geq 1$. Suppose the previous iterate $\mathbf{Z}_{s-1}$ is sufficiently close to $\mathbf{Z}^*$, i.e., $D(\mathbf{Z}_{s-1}, \mathbf{Z}^*) \leq \alpha\sqrt{\sigma_r^*}$, where $\alpha$ is defined in Theorem 5.3. In the following discussions, we are going to show that the following contraction result with respect to the $s$-th iterate:

$$D^2(\mathbf{Z}_s, \mathbf{Z}^*) \leq \left(1 - \frac{\eta\sigma_r^*}{16}\right) \cdot D^2(\mathbf{Z}_{s-1}, \mathbf{Z}^*) + 3\delta \cdot \alpha\sqrt{\sigma_r^*} + 2\delta^2 \tag{A.9}$$

holds with probability at least $1 - c_1/d$, where $\mathbf{Z}_s = [\mathbf{U}_s; \mathbf{V}_s]$ and $\mathbf{Z}_{s-1} = [\mathbf{U}_{s-1}; \mathbf{V}_{s-1}]$.

Denote the optimal rotation with respect to $\mathbf{Z}_{\text{init}}$ by $\mathbf{R}_{\text{init}}$ such that $\mathbf{R}_{\text{init}} = \arg\min_{\mathbf{R}\in\mathbb{Q}_r}\|\mathbf{Z}_{\text{init}} - \mathbf{Z}^*\mathbf{R}\|_F$. Since the initial iterate $\mathbf{Z}_{\text{init}}$ satisfies $D(\mathbf{Z}_{\text{init}}, \mathbf{Z}^*) \leq \sqrt{\sigma_r^*}/40$, we have $\|\mathbf{Z}_{\text{init}} - \mathbf{Z}^*\mathbf{R}_{\text{init}}\|_2 \leq \sqrt{\sigma_r^*}/40$, which implies

$$\sqrt{\sigma_1^*} \leq \|\mathbf{Z}^*\mathbf{R}_{\text{init}}\|_2 - \|\mathbf{Z}_{\text{init}} - \mathbf{Z}^*\mathbf{R}_{\text{init}}\|_2 \leq \|\mathbf{Z}_{\text{init}}\|_2 \leq \|\mathbf{Z}^*\mathbf{R}_{\text{init}}\|_2 + \|\mathbf{Z}_{\text{init}} - \mathbf{Z}^*\mathbf{R}_{\text{init}}\|_2 \leq 2\sqrt{\sigma_1^*}.$$

Thus, according to the definition of $\mathcal{C}_1, \mathcal{C}_2$ in (4.2) and Assumption 3.1, we have

$$\|\mathbf{X}_L\mathbf{U}^*\|_{2,\infty} \leq \|\mathbf{X}_L\overline{\mathbf{U}}^*\|_{2,\infty} \cdot \|\mathbf{\Sigma}^*\|_2^{1/2} \leq \sqrt{\frac{\mu r \sigma_1^*}{d_1}} \leq \sqrt{\frac{\mu r}{d_1}} \cdot \|\mathbf{Z}_{\text{init}}\|_2,$$

which implies that $\mathbf{U}^* \in \mathcal{C}_1$. Similarly, we can derive that $\mathbf{V}^* \in \mathcal{C}_2$. In addition, based on the definition of the constraint sets $\mathcal{C}_1$ and $\mathcal{C}_2$ in (4.2), we further have $\|\mathbf{U}\|_{2,\infty} \leq 2\sqrt{\mu_0 r \sigma_1^*/d_1}$ and $\|\mathbf{V}\|_{2,\infty} \leq 2\sqrt{\mu_0 r \sigma_1^*/d_2}$. For any $s \in \{1, 2, \ldots, S\}$, we denote $\mathbf{R}_s = \arg\min_{\mathbf{R}\in\mathbb{Q}_r}\|\mathbf{Z}_s - \mathbf{Z}^*\mathbf{R}\|_F$ as the optimal rotation with respect to $\mathbf{Z}_s$, and we let $\mathbf{H}_s = \mathbf{Z}_s - \mathbf{Z}^*\mathbf{R}_s$. Consider the $s$-iteration of Phase 2 in Algorithm 1, we let $\widehat{\mathbf{U}}_s = \mathbf{U}_{s-1} - \eta\nabla_{\mathbf{U}}f_{\Omega_s}(\mathbf{U}_{s-1}, \mathbf{V}_{s-1})$ and $\widehat{\mathbf{V}}_s = \mathbf{V}_{s-1} - \eta\nabla_{\mathbf{V}}f_{\Omega_s}(\mathbf{U}_{s-1}, \mathbf{V}_{s-1})$. Thus based on the update rule, we have

$$D^2(\mathbf{Z}_s, \mathbf{Z}^*) \leq \|\mathbf{U}_s - \mathbf{U}^*\mathbf{R}_{s-1}\|_F^2 + \|\mathbf{V}_s - \mathbf{V}^*\mathbf{R}_{s-1}\|_F^2$$
$$= \underbrace{\|\mathcal{P}_{\mathcal{C}_1}(\widehat{\mathbf{U}}_s, \delta) - \mathbf{U}^*\mathbf{R}_{s-1}\|_F^2}_{I_1} + \underbrace{\|\mathcal{P}_{\mathcal{C}_2}(\widehat{\mathbf{V}}_s, \delta) - \mathbf{V}^*\mathbf{R}_{s-1}\|_F^2}_{I_2},$$

where the first inequality follows from Definition 5.1, and the second inequality follows from the update rule. As for the first term $I_1$, we have

$$I_1 = \|\mathcal{P}_{\mathcal{C}_1}(\widehat{\mathbf{U}}_s) - \mathbf{U}^*\mathbf{R}_{s-1}\|_F^2 + 2\langle\mathcal{P}_{\mathcal{C}_1}(\widehat{\mathbf{U}}_s, \delta) - \mathcal{P}_{\mathcal{C}_1}(\widehat{\mathbf{U}}_s), \mathcal{P}_{\mathcal{C}_1}(\widehat{\mathbf{U}}_s) - \mathbf{U}^*\mathbf{R}_{s-1}\rangle$$
$$+ \|\mathcal{P}_{\mathcal{C}_1}(\widehat{\mathbf{U}}_s, \delta) - \mathcal{P}_{\mathcal{C}_1}(\widehat{\mathbf{U}}_s)\|_F^2$$
$$\leq \|\mathcal{P}_{\mathcal{C}_1}(\widehat{\mathbf{U}}_s) - \mathbf{U}^*\mathbf{R}_{s-1}\|_F^2 + 2\delta \cdot \|\mathcal{P}_{\mathcal{C}_1}(\widehat{\mathbf{U}}_s) - \mathbf{U}^*\mathbf{R}_{s-1}\|_F + \delta^2$$
$$\leq \|\mathbf{U}_{s-1} - \eta\nabla_{\mathbf{U}}f_{\Omega_s}(\mathbf{U}_{s-1}, \mathbf{V}_{s-1}) - \mathbf{U}^*\mathbf{R}_{s-1}\|_F^2 + 2\delta \cdot \|\mathbf{U}_{s-1} - \eta\nabla_{\mathbf{U}}f_{\Omega_s}(\mathbf{U}_{s-1}, \mathbf{V}_{s-1}) - \mathbf{U}^*\mathbf{R}_{s-1}\|_F + \delta^2,$$

where the first inequality holds because $\mathcal{P}_{\mathcal{C}_1}(\widehat{\mathbf{U}}_s, \delta)$ is the $\delta$-approximate solution and $\mathcal{P}_{\mathcal{C}_1}(\widehat{\mathbf{U}}_s)$ is the exact solution to the same optimization problem, and the second inequality is due to the non-expansive property of projection $\mathcal{P}_{\mathcal{C}_1}$ and the fact that $\mathbf{U}^* \in \mathcal{C}_1$. Based on the similar technique, we can upper bound $I_2$ as follows

$$I_2 \leq \|\mathbf{V}_{s-1} - \eta\nabla_{\mathbf{V}}f_{\Omega_s}(\mathbf{U}_{s-1}, \mathbf{V}_{s-1}) - \mathbf{V}^*\mathbf{R}_{s-1}\|_F^2 + 2\delta \cdot \|\mathbf{V}_{s-1} - \eta\nabla_{\mathbf{V}}f_{\Omega_s}(\mathbf{U}_{s-1}, \mathbf{V}_{s-1}) - \mathbf{V}^*\mathbf{R}_{s-1}\|_F + \delta^2.$$

Therefore, combining the upper bounds of $I_1$ and $I_2$, we have

$$D^2(\mathbf{Z}_s, \mathbf{Z}^*) \leq D^2(\mathbf{Z}_{s-1}, \mathbf{Z}^*) - 2\eta\langle\nabla\widetilde{f}_{\Omega_s}(\mathbf{Z}_{s-1}), \mathbf{H}_{s-1}\rangle + \eta^2\|\nabla\widetilde{f}_{\Omega_s}(\mathbf{Z}_{s-1})\|_F^2$$
$$+ 2\sqrt{2}\delta \cdot \left(D^2(\mathbf{Z}_{s-1}, \mathbf{Z}^*) - 2\eta\langle\nabla\widetilde{f}_{\Omega_s}(\mathbf{Z}_{s-1}), \mathbf{H}_{s-1}\rangle + \eta^2\|\nabla\widetilde{f}_{\Omega_s}(\mathbf{Z}_{s-1})\|_F^2\right)^{1/2} + 2\delta^2,$$

where $\widetilde{f}_{\Omega_s}$ is defined in (A.3), and the inequality follows from the triangle inequality.

It is worth noting that the subsampling technique ensures the previous iterate $\mathbf{Z}_{s-1}$ is independent of the samples $\Omega_s$ used in the $s$-th iteration. According to the assumptions of Theorem 5.3, $|\Omega_s| = |\Omega|/(2S) \leq c_1 \max\{\mu_0^2 r^2 \kappa^2, \mu_0 \mu_1 r \kappa n\} \log d$, thus we can directly apply Lemmas A.1 and A.2. More specifically, with probability at least $1 - c_1/d$, we have

$$\langle \nabla \widetilde{f}_{\Omega_s}(\mathbf{Z}_{s-1}), \mathbf{H}_{s-1} \rangle \geq \frac{1}{20} \|\mathbf{Z}_{s-1}\mathbf{Z}_{s-1}^\top - \mathbf{Z}^*\mathbf{Z}^{*\top}\|_F^2 + \frac{1}{4\sigma_1^*} \|\widetilde{\mathbf{Z}}^* \widetilde{\mathbf{Z}}^{*\top} \mathbf{Z}_{s-1}\|_F^2 + \frac{\sigma_r^*}{8} \|\mathbf{H}_{s-1}\|_F^2 - 40 \|\mathbf{H}_{s-1}\|_F^4,$$

where $\widetilde{\mathbf{Z}}^* = [\mathbf{U}^*; -\mathbf{V}^*]$. In addition, we have

$$\|\nabla \widetilde{f}_{\Omega_s}(\mathbf{Z}_s)\|_F^2 \leq (16r + 4)\sigma_1^* \|\mathbf{Z}_{s-1}\mathbf{Z}_{s-1}^\top - \mathbf{Z}^*\mathbf{Z}^{*\top}\|_F^2 + \|\widetilde{\mathbf{Z}}^* \widetilde{\mathbf{Z}}^{*\top} \mathbf{Z}_{s-1}\|_F^2 + 4r\sigma_r^{*2} \|\mathbf{H}_{s-1}\|_F^2.$$

Thus, by setting step size $\eta \leq 1/(200r\sigma_1^*)$, we obtain

$$-2\eta \langle \nabla \widetilde{f}_{\Omega_s}(\mathbf{Z}_{s-1}), \mathbf{H}_{s-1} \rangle + \eta^2 \|\nabla \widetilde{f}_{\Omega_s}(\mathbf{Z}_s)\|_F^2 \leq -\frac{\eta \sigma_r^*}{8} \|\mathbf{H}_{s-1}\|_F^2 + 80\eta \|\mathbf{H}_{s-1}\|_F^4,$$

which implies that the induction hypothesis that $D(\mathbf{Z}_{s-1}, \mathbf{Z}^*) = \|\mathbf{H}_{s-1}\|_F \leq \alpha \sqrt{\sigma_r^*}$, we have

$$D^2(\mathbf{Z}_s, \mathbf{Z}^*) \leq \left(1 - \frac{\eta \sigma_r^*}{16}\right) \cdot D^2(\mathbf{Z}_{s-1}, \mathbf{Z}^*) + 2\delta \cdot \sqrt{2\left(1 - \frac{\eta \sigma_r^*}{16}\right)} \cdot D(\mathbf{Z}_{s-1}, \mathbf{Z}^*) + 2\delta^2$$

$$\leq \left(1 - \frac{\eta \sigma_r^*}{16}\right) \cdot D^2(\mathbf{Z}_{s-1}, \mathbf{Z}^*) + 3\delta \cdot \alpha \sqrt{\sigma_r^*} + 2\delta^2,$$

which completes the proof of (A.9).

Moreover, according to the assumption that $D(\mathbf{Z}_{\text{init}}, \mathbf{Z}^*) \leq \alpha \sqrt{\sigma_r^*}/2$, if we choose $\delta \leq \alpha \sqrt{\sigma_r^*}/(2\sqrt{2})$, then we have

$$D^2(\mathbf{Z}_0, \mathbf{Z}^*) \leq \|\mathbf{U}_0 - \mathbf{U}^* \mathbf{R}_{\text{init}}\|_F^2 + \|\mathbf{V}_0 - \mathbf{V}^* \mathbf{R}_{\text{init}}\|_F^2$$

$$\leq \|\mathcal{P}_{\mathcal{C}_1}(\mathbf{U}_{\text{init}}) - \mathbf{U}^* \mathbf{R}_{\text{init}}\|_F^2 + \|\mathcal{P}_{\mathcal{C}_2}(\mathbf{V}_{\text{init}}) - \mathbf{V}^* \mathbf{R}_{\text{init}}\|_F^2$$

$$+ 2\delta \cdot \left(\|\mathcal{P}_{\mathcal{C}_1}(\mathbf{U}_{\text{init}}) - \mathbf{U}^* \mathbf{R}_{\text{init}}\|_F + \|\mathcal{P}_{\mathcal{C}_2}(\mathbf{V}_{\text{init}}) - \mathbf{V}^* \mathbf{R}_{\text{init}}\|_F\right) + 2\delta^2$$

$$\leq D^2(\mathbf{Z}_{\text{init}}, \mathbf{Z}^*) + 2\sqrt{2}\delta \cdot D(\mathbf{Z}_{\text{init}}, \mathbf{Z}^*) + 2\delta^2 \leq \alpha^2 \sigma_r^*,$$

where we let $\mathbf{R}_{\text{init}}$ be the optimal rotation with respect to $\mathbf{Z}_{\text{init}}$, and the last inequality follows from the non-expansiveness property of projection onto convex set. Thus, we have shown that the induction hypothesis $D(\mathbf{Z}_{s-1}, \mathbf{Z}^*) \leq \alpha \sqrt{\sigma_r^*}$ holds for the first iterate.

To this end, it remains to verify the induction step, or more specifically, $D(\mathbf{Z}_{s-1}, \mathbf{Z}^*) \leq \alpha \sqrt{\sigma_r^*}$ implies $D(\mathbf{Z}_s, \mathbf{Z}^*) \leq \alpha \sqrt{\sigma_r^*}$, for any $s \geq 1$. This step can be proved based on (A.9): with high probability, we have

$$D^2(\mathbf{Z}_s, \mathbf{Z}^*) \leq \left(1 - \frac{\eta \sigma_r^*}{16}\right) \cdot D^2(\mathbf{Z}_{s-1}, \mathbf{Z}^*) + 3\delta \cdot \alpha \sqrt{\sigma_r^*} + 2\delta^2$$

$$\leq \left(1 - \frac{\eta \sigma_r^*}{16}\right) \cdot \alpha^2 \sigma_r^* + 3\delta \cdot \alpha \sqrt{\sigma_r^*} + 2\delta^2 \leq \alpha^2 \sigma_r^*,$$

provided that $\delta \leq c_2 \sqrt{\sigma_r^*}/(r\kappa)$ with constant $c_2$ sufficiently small. Finally, by induction and union bound, we obtain

$$D^2(\mathbf{Z}_S, \mathbf{Z}^*) \leq \left(1 - \frac{\eta \sigma_r^*}{16}\right)^S \cdot D^2(\mathbf{Z}_0, \mathbf{Z}^*) + \frac{16}{\eta \sigma_r^*} \cdot (3\delta \cdot \alpha \sqrt{\sigma_r^*} + 2\delta^2)$$

holds with probability at least $1 - c_1 S/d$, which completes the proof. □

## A.3 Proof of Theorem 5.4

The proof of Theorem 5.4 will be similar to the proof of Theorem 5.3. The only difference is that we do not require sample splitting in Phase 3, thus the iterates are no longer independent from the subset of samples. The following curvature and smoothness of $\widetilde{f}_\Omega$ are proved for all $\mathbf{Z} \in \mathbb{R}^{(n_1+n_2) \times r}$ satisfying $D(\mathbf{Z}, \mathbf{Z}^*) \leq c_0 \sqrt{\sigma_r^*}/(\mu_1 n)$. The proofs are presented in Sections B.3 and B.4 respectively.

**Lemma A.3** (Local curvature). Under the previously stated assumptions in Theorem 5.4, for all $\mathbf{Z} = [\mathbf{U}; \mathbf{V}] \in \mathbb{R}^{(n_1+n_2) \times r}$ such that $D(\mathbf{Z}, \mathbf{Z}^*) \leq c_0 \sqrt{\sigma_r^*}/(\mu_1 n)$ with constant $c_0$ small enough, there exists constants $c_1, c_2$ such that if $|\Omega| \geq c_1 \mu_0 \mu_1 rn \log d$, then with probability at least $1 - c_2/d$, we have

$$\langle \nabla \widetilde{f}_\Omega(\mathbf{Z}), \mathbf{H} \rangle \geq \frac{1}{20} \|\mathbf{Z}\mathbf{Z}^\top - \mathbf{Z}^*\mathbf{Z}^{*\top}\|_F^2 + \frac{1}{4\sigma_1^*} \|\widetilde{\mathbf{Z}}^* \widetilde{\mathbf{Z}}^{*\top} \mathbf{Z}\|_F^2 + \frac{\sigma_r^*}{8} \|\mathbf{H}\|_F^2 - 20\|\mathbf{H}\|_F^4,$$

where $\widetilde{\mathbf{Z}}^* = [\mathbf{U}^*; -\mathbf{V}^*]$.

**Lemma A.4** (Local smoothness). Under the previous stated assumptions as in Theorem 5.4, for all $\mathbf{Z} = [\mathbf{U}; \mathbf{V}] \in \mathbb{R}^{(n_1+n_2) \times r}$ such that $D(\mathbf{Z}, \mathbf{Z}^*) \leq c_0 \sqrt{\sigma_r^*}/(\mu_1 n)$ with constant $c_0$ small enough, there exist constants $c_1, c_2$ such that if $|\Omega| \geq c_1 \mu_0 \mu_1 rn \log d$, then with probability at least $1 - c_2/d$, we have

$$\|\nabla \widetilde{f}_\Omega(\mathbf{Z})\|_F^2 \leq 84\sigma_1^* \|\mathbf{Z}\mathbf{Z}^\top - \mathbf{Z}^*\mathbf{Z}^{*\top}\|_F^2 + \|\widetilde{\mathbf{Z}}^* \widetilde{\mathbf{Z}}^{*\top} \mathbf{Z}\|_F^2 + 4\sigma_1^* \sigma_r^* \|\mathbf{H}\|_F^2 + 40\sigma_1^* \|\mathbf{H}\|_F^4,$$

where $\widetilde{\mathbf{Z}}^* = [\mathbf{U}^*; -\mathbf{V}^*]$.

Now we are ready to prove Theorem 5.4.

*Proof.* For any $t \in \{0, 1, \ldots, T\}$, we denote $\mathbf{R}^t = \operatorname{argmin}_{\mathbf{R} \in \mathbb{Q}_r} \|\mathbf{Z}^t - \mathbf{Z}^* \mathbf{R}\|_F$ as the optimal rotation with respect to $\mathbf{Z}^t$, and we let $\mathbf{H}^t = \mathbf{Z}^t - \mathbf{Z}^* \mathbf{R}^t$. Note that the initial iterate of Phase 3 in Algorithm 1 satisfying $D(\mathbf{Z}^0, \mathbf{Z}^*) \leq c_0 \sqrt{\sigma_r^*}/(\mu n)$. Assume the induction hypothesis $D(\mathbf{Z}^s, \mathbf{Z}^*) \leq c_0 \sqrt{\sigma_r^*}/(\mu n)$ holds for $s = 1, 2, \ldots, t$. Consider the $t$-th iteration, based on the update rule, we have

$$\begin{aligned}
D^2(\mathbf{Z}^{t+1}, \mathbf{Z}^*) &\leq \|\mathbf{U}^{t+1} - \mathbf{U}^* \mathbf{R}^t\|_F^2 + \|\mathbf{V}^{t+1} - \mathbf{V}^* \mathbf{R}^t\|_F^2 \\
&= \|\mathbf{U}^t - \tau \nabla_\mathbf{U} f_\Omega(\mathbf{U}^t, \mathbf{V}^t) - \mathbf{U}^* \mathbf{R}^t\|_F^2 + \|\mathbf{V}^t - \tau \nabla_\mathbf{V} f_\Omega(\mathbf{U}^t, \mathbf{V}^t) - \mathbf{V}^* \mathbf{R}^t\|_F^2 \\
&= D^2(\mathbf{Z}^t, \mathbf{Z}^*) - 2\tau \langle \nabla \widetilde{f}_\Omega(\mathbf{Z}^t), \mathbf{H}^t \rangle + \tau^2 \|\nabla \widetilde{f}_\Omega(\mathbf{Z}^t)\|_F^2,
\end{aligned}$$

where the inequality follows from Definition 5.1. According to the assumptions of Theorem 5.4, we can directly apply Lemmas A.3 and A.4. More specifically, with probability at least $1 - c'/d$, we have

$$\langle \nabla \widetilde{f}_\Omega(\mathbf{Z}^t), \mathbf{H}^t \rangle \geq \frac{1}{20} \|\mathbf{Z}^t \mathbf{Z}^{t\top} - \mathbf{Z}^* \mathbf{Z}^{*\top}\|_F^2 + \frac{1}{4\sigma_1^*} \|\widetilde{\mathbf{Z}}^* \widetilde{\mathbf{Z}}^{*\top} \mathbf{Z}^t\|_F^2 + \frac{\sigma_r^*}{8} \|\mathbf{H}^t\|_F^2 - 20 \|\mathbf{H}^t\|_F^4,$$

where $\widetilde{\mathbf{Z}}^* = [\mathbf{U}^*; -\mathbf{V}^*]$. In addition, we have

$$\|\nabla \widetilde{f}_\Omega(\mathbf{Z}^t)\|_F^2 \leq 84\sigma_1^* \|\mathbf{Z}^t \mathbf{Z}^{t\top} - \mathbf{Z}^* \mathbf{Z}^{*\top}\|_F^2 + \|\widetilde{\mathbf{Z}}^* \widetilde{\mathbf{Z}}^{*\top} \mathbf{Z}^t\|_F^2 + 4\sigma_1^* \sigma_r^* \|\mathbf{H}^t\|_F^2 + 40\sigma_1^* \|\mathbf{H}^t\|_F^4.$$

Thus, by setting step size $\tau \leq c_1/\sigma_1^*$ with $c_1 \leq 1/840$, we obtain

$$-2\tau \langle \nabla \widetilde{f}_{\Omega_s}(\mathbf{Z}^t), \mathbf{H}^t \rangle + \tau^2 \|\nabla \widetilde{f}_\Omega(\mathbf{Z}^t)\|_F^2 \leq -\frac{\eta \sigma_r^*}{8} \|\mathbf{H}^t\|_F^2 + 50\eta \|\mathbf{H}^t\|_F^4,$$

which implies that under condition that $D(\mathbf{Z}^t, \mathbf{Z}^*) = \|\mathbf{H}^t\|_F \leq c_2 \sqrt{\sigma_r^*}$ with $c_2 \leq 1/30$, we have

$$D^2(\mathbf{Z}^{t+1}, \mathbf{Z}^*) \leq \left(1 - \frac{\tau \sigma_r^*}{16}\right) \cdot D^2(\mathbf{Z}^t, \mathbf{Z}^*),$$

which implies the $(t+1)$-th iterate $\mathbf{Z}^{t+1}$ still satisfies $D^2(\mathbf{Z}^{t+1}, \mathbf{Z}^*) \leq c_0 \sqrt{\sigma_r^*}/(\mu n)$. Thus by induction, we complete the proof. □

## A.4 Proof of Theorem 5.5

*Proof.* To prove the overall theoretical guarantee of Algorithm 1, we need to examine the conditions required by previous main theorems regarding the three phases. More specifically, to ensure the $O(\sqrt{\sigma_r^*})$ initial ball assumption $D(\mathbf{Z}_{\text{init}}, \mathbf{Z}^*) \leq \alpha\sqrt{\sigma_r^*}/2$ in Theorem 5.3, it suffices to set $\delta \leq \alpha/8$ in Theorem 5.2, which implies the sample complexity required by Phase 1 is $|\Omega_0| = O(r^2\kappa^2 n \log d)$. In addition, according to Theorem 5.3, we have

$$D^2(\mathbf{Z}_S, \mathbf{Z}^*) \leq \left(1 - \frac{c_2}{16r\kappa}\right)^S \cdot \alpha^2 \sigma_r^* + c_4 \delta \cdot r\kappa\sqrt{\sigma_r^*}.$$

Thus, in order to guarantee the $O(\sqrt{\sigma_r^*}/n)$ initial ball assumption $D(\mathbf{Z}_S, \mathbf{Z}^*) \leq c_0\sqrt{\sigma_r^*}/(\mu_1 n)$ holds in Theorem 5.4, it is sufficient to perform $S = O(r\kappa \log n)$ iterations in Phase 2 of Algorithm 1 and let the approximation error $\delta = O(1/(r\kappa n^2))$. Based on Theorem 5.3, we derive the sample complexity required by Phase 2 is $c \cdot \max\{\mu_1 n, \mu_0 r\kappa\} \mu_0 r^2 \kappa^2 \log n \log d$. Together with Theorem 5.4, we conclude that the overall sample complexity of Algorithm 1.

Finally, as for the reconstruction error $\|\mathbf{M}^T - \mathbf{M}^*\|_F$, let $\mathbf{R}^T$ be the optimal rotation between $\mathbf{Z}^T$ and $\mathbf{Z}^*$, then we have

$$\begin{aligned}
\|\mathbf{U}^T(\mathbf{V}^T)^\top - \mathbf{U}^*\mathbf{V}^{*\top}\|_F &\leq \|\mathbf{U}^T(\mathbf{V}^T - \mathbf{V}^*\mathbf{R}^T)^\top\|_F + \|(\mathbf{U}^T - \mathbf{U}^*\mathbf{R}^T)(\mathbf{V}^*\mathbf{R}^T)^\top\|_F \\
&\leq \|\mathbf{U}^T\|_2 \cdot \|\mathbf{V}^T - \mathbf{V}^*\mathbf{R}^T\|_F + \|\mathbf{V}^*\mathbf{R}^T\|_2 \cdot \|\mathbf{U}^T - \mathbf{U}^*\mathbf{R}^T\|_F \\
&\leq 3\sqrt{\sigma_1^*} \cdot D(\mathbf{Z}^T, \mathbf{Z}^*),
\end{aligned} \quad (A.10)$$

where the second inequality is due to $\|\mathbf{AB}\|_F \leq \|\mathbf{A}\|_2 \cdot \|\mathbf{B}\|_F$, and the last inequality follows from the fact that $\|\mathbf{Z}^T - \mathbf{Z}^*\mathbf{R}^T\|_F \leq D(\mathbf{Z}_S, \mathbf{Z}^*) \leq \alpha\sqrt{\sigma_r^*}/(\mu_1 n) \leq \sqrt{\sigma_1^*}$. Noticing that according to Theorem 5.4, Phase 3 achieves linear rate of convergence, which implies that with $T = O(\kappa \log(1/\epsilon))$ iterations, we have $D(\mathbf{Z}^T, \mathbf{Z}^*) \leq \epsilon$. Combining with (A.10), we complete the proof. $\square$

# B Proofs of the Technical Lemmas in Section A

In this section, we provide the theoretical proofs of the technical lemmas used in Section A.

## B.1 Proof of Lemma A.1

To prove Lemma A.1, we need to make use of the following auxiliary lemmas. Inspired by Tu et al. (2015), we show that the reference function $G(\mathbf{Z})$ has a similar local curvature property in Lemma B.1. Based on Matrix Bernstein Inequality, Lemma B.2 generalizes the results of Theorem 4.1 in Candès and Recht (2009) to inductive setting, while Lemma B.3 provides the high probability bound on the remaining term.

**Lemma B.1.** Let $\mathbf{Z}, \mathbf{Z}^* \in \mathbb{R}^{(n_1+n_2) \times r}$, and $G(\mathbf{Z}) = \|\mathbf{Z}\mathbf{Z}^\top - \mathbf{Z}^*\mathbf{Z}^{*\top}\|_F^2/4$. For any $\mathbf{Z}$ satisfying

$$\|\mathbf{Z} - \mathbf{Z}^*\mathbf{R}\|_2^2 \leq \sigma_r^2(\mathbf{Z}^*)/5, \text{ where } \mathbf{R} = \operatorname*{argmin}_{\widetilde{\mathbf{R}} \in \mathbb{Q}_r} \|\mathbf{Z} - \mathbf{Z}^*\widetilde{\mathbf{R}}\|_F,$$

we have

$$\langle \nabla G(\mathbf{Z}), \mathbf{Z} - \mathbf{Z}^*\mathbf{R} \rangle \geq \frac{\sigma_r^2(\mathbf{Z}^*)}{4}\|\mathbf{Z} - \mathbf{Z}^*\mathbf{R}\|_F^2 + \frac{1}{4}\|\mathbf{Z}\mathbf{Z}^\top - \mathbf{Z}^*\mathbf{Z}^{*\top}\|_F^2.$$

**Lemma B.2.** Assume the previously stated assumptions in Lemma A.1 hold. Define subspace

$$\mathcal{T} = \left\{\overline{\mathbf{U}}^*\mathbf{A}^\top + \mathbf{B}\overline{\mathbf{V}}^{*\top}, \text{ for some } \mathbf{A} \in \mathbb{R}^{n_2 \times r} \text{ and } \mathbf{B} \in \mathbb{R}^{n_1 \times r}\right\}.$$

Let $\mathcal{P}_{\mathcal{T}} : \mathbb{R}^{n_1 \times n_2} \to \mathbb{R}^{n_1 \times n_2}$ be the Euclidean projection onto $\mathcal{T}$. Specifically, for any $\mathbf{Z} \in \mathbb{R}^{n_1 \times n_2}$, we have

$$\mathcal{P}_{\mathcal{T}}(\mathbf{Z}) = \overline{\mathbf{U}}^* \overline{\mathbf{U}}^{*\top} \mathbf{Z} + \mathbf{Z} \overline{\mathbf{V}}^* \overline{\mathbf{V}}^{*\top} - \overline{\mathbf{U}}^* \overline{\mathbf{U}}^{*\top} \mathbf{Z} \overline{\mathbf{V}}^* \overline{\mathbf{V}}^{*\top}. \tag{B.1}$$

For any $\gamma \in (0, 1)$, there exist constants $c_1$, $c_2$ such that if $|\Omega| \geq c_1 \mu_0 \mu_1 r n \log d / \gamma^2$, then for all $\mathbf{Z} \in \mathbb{R}^{n_1 \times n_2}$, with probability at least $1 - c_2/d$, we have

$$\left\| \mathcal{P}_{\mathcal{T}}(\mathbf{Z}) - p^{-1} \mathcal{P}_{\mathcal{T}} \left( \mathbf{X}_L^\top \mathcal{P}_\Omega (\mathbf{X}_L \mathcal{P}_{\mathcal{T}}(\mathbf{Z}) \mathbf{X}_R^\top) \mathbf{X}_R \right) \right\|_F \leq \gamma \|\mathbf{Z}\|_F. \tag{B.2}$$

Moreover for all $\mathbf{Z}_1, \mathbf{Z}_2 \in \mathcal{T}$, we have

$$\left| \langle \mathbf{X}_L \mathbf{Z}_1 \mathbf{X}_R^\top - p^{-1} \mathcal{P}_\Omega (\mathbf{X}_L \mathbf{Z}_1 \mathbf{X}_R^\top), \mathbf{X}_L \mathbf{Z}_2 \mathbf{X}_R^\top \rangle \right| \leq \gamma \|\mathbf{Z}_1\|_F \cdot \|\mathbf{Z}_2\|_F,$$

and for all $\mathbf{Z} \in \mathcal{T}$, we have

$$p^{-1} \left\| \mathcal{P}_\Omega (\mathbf{X}_L \mathbf{Z} \mathbf{X}_R^\top) \right\|_F^2 \leq (1 + \gamma) \|\mathbf{Z}\|_F^2.$$

**Lemma B.3.** Assume the previously stated assumptions in Lemma A.1 hold. For any fixed $\mathbf{U} \in \mathbb{R}^{n_1 \times r}$, $\mathbf{V} \in \mathbb{R}^{n_2 \times r}$ satisfying $\|\mathbf{X}_L \mathbf{U}\|_{2,\infty} \leq 3\sqrt{\mu_0 r \sigma_1^*/d_1}$ and $\|\mathbf{X}_R \mathbf{V}\|_{2,\infty} \leq 3\sqrt{\mu_0 r \sigma_1^*/d_2}$, there exist constants $c_1, c_2$ such that with probability $1 - c_1/d$, we have

$$\frac{1}{p} \left\| \mathcal{P}_\Omega (\mathbf{X}_L \mathbf{U} \mathbf{V}^\top \mathbf{X}_R^\top) \right\|_F^2 \leq \|\mathbf{U} \mathbf{V}^\top\|_F^2 + \gamma \sigma_r^* \cdot (\|\mathbf{U}\|_F^2 + \|\mathbf{V}\|_F^2),$$

provided that $|\Omega| \geq c_2 \max\{\mu_0^2 r^2 \kappa^2, \mu_0 \mu_1 r \kappa n\} \log d / \gamma^2$.

Now we are ready to prove Lemma A.1.

*Proof of Lemma A.1.* Recall that $\mathbf{R} = \arg\min_{\widetilde{\mathbf{R}} \in \mathbb{Q}_r} \|\mathbf{Z} - \mathbf{Z}^* \widetilde{\mathbf{R}}\|_F$ and $\mathbf{H} = \mathbf{Z} - \mathbf{Z}^* \mathbf{R}$. According to the gradient of $\widetilde{f}_\Omega$ in (A.8), we have

$$\langle \nabla \widetilde{f}_\Omega(\mathbf{Z}), \mathbf{H} \rangle = \underbrace{\frac{1}{2} \langle \nabla G(\mathbf{Z}), \mathbf{H} \rangle + \frac{1}{2} \langle \widetilde{\mathbf{Z}}^* \widetilde{\mathbf{Z}}^{*\top} \mathbf{Z}, \mathbf{H} \rangle}_{I_1} + \underbrace{\langle (p^{-1} \mathcal{P}_{\widetilde{\Omega}} - \mathcal{P}_{\text{off}})(\mathbf{X}[\mathbf{Z}\mathbf{Z}^\top - \mathbf{Z}^* \mathbf{Z}^{*\top}]\mathbf{X}^\top), \mathbf{X} \mathbf{H} (\mathbf{X} \mathbf{Z})^\top \rangle}_{I_2}.$$

In the sequel, we are going to bound the terms $I_1$ and $I_2$, respectively.

**Lower bound of $I_1$.** According to the assumption $\|\mathbf{H}\|_F \leq \sqrt{2\sigma_r^*/5}$, we have $\|\mathbf{Z} - \mathbf{Z}^* \mathbf{R}\|_2^2 \leq \sigma_r^2(\mathbf{Z}^*)/5$. Thus, we can apply Lemma B.1 directly

$$\begin{aligned} I_1 &= \frac{1}{2} \langle \nabla G(\mathbf{Z}), \mathbf{H} \rangle + \frac{1}{2} \operatorname{tr}(\mathbf{Z}^\top \widetilde{\mathbf{Z}}^* \widetilde{\mathbf{Z}}^{*\top} \mathbf{Z}) \\ &\geq \frac{\sigma_r^2(\mathbf{Z}^*)}{8} \|\mathbf{H}\|_F^2 + \frac{1}{8} \|\mathbf{Z}\mathbf{Z}^\top - \mathbf{Z}^* \mathbf{Z}^{*\top}\|_F^2 + \frac{1}{2\|\widetilde{\mathbf{Z}}^*\|_2^2} \|\widetilde{\mathbf{Z}}^* \widetilde{\mathbf{Z}}^{*\top} \mathbf{Z}\|_F^2 \\ &= \frac{\sigma_r^*}{4} \|\mathbf{H}\|_F^2 + \frac{1}{8} \|\mathbf{Z}\mathbf{Z}^\top - \mathbf{Z}^* \mathbf{Z}^{*\top}\|_F^2 + \frac{1}{4\sigma_1^*} \|\widetilde{\mathbf{Z}}^* \widetilde{\mathbf{Z}}^{*\top} \mathbf{Z}\|_F^2, \end{aligned} \tag{B.3}$$

where the first equality holds because $\mathbf{Z}^{*\top} \widetilde{\mathbf{Z}}^* = \mathbf{0}$, the second inequality follows from Lemma B.1 and the fact that $\|\widetilde{\mathbf{Z}}^* \widetilde{\mathbf{Z}}^{*\top} \mathbf{Z}\|_F^2 \leq \|\widetilde{\mathbf{Z}}^*\|_2^2 \cdot \|\widetilde{\mathbf{Z}}^{*\top} \mathbf{Z}\|_F^2$, and the third equality holds because $\sigma_r^2(\mathbf{Z}^*) = 2\sigma_r^*$ and $\|\widetilde{\mathbf{Z}}^*\|_2^2 = 2\sigma_1^*$.

**Upper bound of $|I_2|$.** Note that $\mathbf{Z} = [\mathbf{U}; \mathbf{V}]$ and $\mathbf{M}^* = \mathbf{U}^* \mathbf{V}^{*\top}$, we denote $\mathbf{M} = \mathbf{U} \mathbf{V}^\top$, $\widetilde{\mathbf{U}} = \mathbf{U}^* \mathbf{R}$ and $\widetilde{\mathbf{V}} = \mathbf{V}^* \mathbf{R}$. Besides, let $\mathbf{H}_U \in \mathbb{R}^{n_1 \times r}$, $\mathbf{H}_V \in \mathbb{R}^{n_2 \times r}$ be the top $n_1$ and bottom $n_2$ rows of $\mathbf{H}$, respectively, then we have $\mathbf{U} = \widetilde{\mathbf{U}} + \mathbf{H}_U$ and $\mathbf{V} = \widetilde{\mathbf{V}} + \mathbf{H}_V$. Note that $\mathbf{U}\mathbf{V}^\top - \widetilde{\mathbf{U}}\widetilde{\mathbf{V}}^\top = \widetilde{\mathbf{U}}\mathbf{H}_V^\top + \mathbf{H}_U\widetilde{\mathbf{V}}^\top + \mathbf{H}_U\mathbf{H}_V^\top$, and $\mathbf{H}_U\mathbf{V}^\top + \mathbf{U}\mathbf{H}_V^\top = \widetilde{\mathbf{U}}\mathbf{H}_V^\top + \mathbf{H}_U\widetilde{\mathbf{V}}^\top + 2\mathbf{H}_U\mathbf{H}_V^\top$. Based on the above notations, we can reformulate $I_2$ as follows

$$
\begin{aligned}
I_2 &= \langle (p^{-1}\mathcal{P}_\Omega - \mathcal{I})(\mathbf{X}_L[\mathbf{U}\mathbf{V}^\top - \widetilde{\mathbf{U}}\widetilde{\mathbf{V}}^\top]\mathbf{X}_R^\top), \mathbf{X}_L(\mathbf{H}_U\mathbf{V}^\top + \mathbf{U}\mathbf{H}_V^\top)\mathbf{X}_R^\top \rangle \\
&= \underbrace{\langle (p^{-1}\mathcal{P}_\Omega - \mathcal{I})(\mathbf{X}_L[\widetilde{\mathbf{U}}\mathbf{H}_V^\top + \mathbf{H}_U\widetilde{\mathbf{V}}^\top]\mathbf{X}_R^\top), \mathbf{X}_L(\widetilde{\mathbf{U}}\mathbf{H}_V^\top + \mathbf{H}_U\widetilde{\mathbf{V}}^\top)\mathbf{X}_R^\top \rangle}_{I_{21}} \\
&\quad + 3\underbrace{\langle (p^{-1}\mathcal{P}_\Omega - \mathcal{I})(\mathbf{X}_L[\widetilde{\mathbf{U}}\mathbf{H}_V^\top + \mathbf{H}_U\widetilde{\mathbf{V}}^\top]\mathbf{X}_R^\top), \mathbf{X}_L\mathbf{H}_U\mathbf{H}_V^\top\mathbf{X}_R^\top \rangle}_{I_{22}} \\
&\quad + 2\underbrace{\langle (p^{-1}\mathcal{P}_\Omega - \mathcal{I})(\mathbf{X}_L\mathbf{H}_U\mathbf{H}_V^\top\mathbf{X}_R^\top), \mathbf{X}_L\mathbf{H}_U\mathbf{H}_V^\top\mathbf{X}_R^\top \rangle}_{I_{23}}.
\end{aligned}
$$

Note that $\widetilde{\mathbf{U}}\mathbf{H}_V^\top + \mathbf{H}_U\widetilde{\mathbf{V}}^\top$ falls into the subspace $\mathcal{T}$ defined in Lemma B.2. Thus according to Lemma B.2, we obtain

$$|I_{21}| \leq \gamma \|\widetilde{\mathbf{U}}\mathbf{H}_V^\top + \mathbf{H}_U\widetilde{\mathbf{V}}^\top\|_F^2 \leq 2\gamma\|\mathbf{M} - \mathbf{M}^*\|_F^2 + \frac{\gamma}{2}\|\mathbf{H}\|_F^4 \tag{B.4}$$

holds with probability at least $1 - c_2/d$, provided that $|\Omega| \geq c_1\mu_0\mu_1 rn\log d/\gamma^2$, where $\gamma$ is a constant such that $\gamma \in (0,1)$. Here, the second inequality holds because $\|\mathbf{A}+\mathbf{B}\|_F^2 \leq 2\|\mathbf{A}\|_F^2 + 2\|\mathbf{B}\|_F^2$ and $\|\mathbf{H}_U\mathbf{H}_V^\top\|_F \leq \|\mathbf{H}\|_F^2/2$. As for the term $I_{22}$, we have

$$
\begin{aligned}
|I_{22}| &\leq \frac{1}{p}|\langle \mathcal{P}_\Omega(\mathbf{X}_L[\widetilde{\mathbf{U}}\mathbf{H}_V^\top + \mathbf{H}_U\widetilde{\mathbf{V}}^\top]\mathbf{X}_R^\top), \mathbf{X}_L\mathbf{H}_U\mathbf{H}_V^\top\mathbf{X}_R^\top\rangle| + |\langle \mathbf{X}_L(\widetilde{\mathbf{U}}\mathbf{H}_V^\top + \mathbf{H}_U\widetilde{\mathbf{V}}^\top)\mathbf{X}_R^\top, \mathbf{X}_L\mathbf{H}_U\mathbf{H}_V^\top\mathbf{X}_R^\top\rangle| \\
&\leq \frac{1}{2\beta}\cdot p^{-1}\|\mathcal{P}_\Omega(\mathbf{X}_L[\widetilde{\mathbf{U}}\mathbf{H}_V^\top + \mathbf{H}_U\widetilde{\mathbf{V}}^\top]\mathbf{X}_R^\top)\|_F^2 + \frac{\beta}{2}\cdot p^{-1}\|\mathcal{P}_\Omega(\mathbf{X}_L\mathbf{H}_U\mathbf{H}_V^\top\mathbf{X}_R^\top)\|_F^2 \\
&\quad + \frac{1}{2\beta}\|\widetilde{\mathbf{U}}\mathbf{H}_V^\top + \mathbf{H}_U\widetilde{\mathbf{V}}^\top\|_F^2 + \frac{\beta}{2}\|\mathbf{H}_U\mathbf{H}_V^\top\|_F^2 \\
&\leq \frac{\beta}{2}\cdot p^{-1}\|\mathcal{P}_\Omega(\mathbf{X}_L\mathbf{H}_U\mathbf{H}_V^\top\mathbf{X}_R^\top)\|_F^2 + \frac{2+\gamma}{2\beta}\|\widetilde{\mathbf{U}}\mathbf{H}_V^\top + \mathbf{H}_U\widetilde{\mathbf{V}}^\top\|_F^2 + \frac{\beta}{8}\|\mathbf{H}\|_F^4,
\end{aligned} \tag{B.5}
$$

where constant $\beta > 0$ will be specified later. Here, the second inequality follows from the Young's Inequality, and the last inequality is due to Lemma B.2 and the fact that $\|\mathbf{H}_U\mathbf{H}_V^\top\|_F \leq \|\mathbf{H}_U\|_F \cdot \|\mathbf{H}_V\|_F \leq \|\mathbf{H}\|_F^2/2$. Since we have $\|\mathbf{X}_L\mathbf{H}_U\|_{2,\infty} \leq \|\mathbf{X}_L\mathbf{U}\|_{2,\infty} + \|\mathbf{X}_L\widetilde{\mathbf{U}}\|_{2,\infty} \leq 3\sqrt{\mu_0 r\sigma_1^*/d_1}$, and similarly we have $\|\mathbf{X}_R\mathbf{H}_V\|_{2,\infty} \leq 3\sqrt{\mu_0 r\sigma_1^*/d_2}$, according to Lemma B.3, we further obtain

$$\frac{1}{p}\|\mathcal{P}_\Omega(\mathbf{X}_L\mathbf{H}_U\mathbf{H}_V^\top\mathbf{X}_R^\top)\|_F^2 \leq \|\mathbf{H}_U\mathbf{H}_V\|_F^2 + \gamma\sigma_r^*\|\mathbf{H}\|_F^2 \leq \frac{1}{4}\|\mathbf{H}\|_F^4 + \gamma\sigma_r^*\|\mathbf{H}\|_F^2 \tag{B.6}$$

holds with probability at least $1 - c_1/d$, provided that $|\Omega| \geq c_2\max\{\mu_0^2 r^2\kappa^2, \mu_0\mu_1 r\kappa n\}\log d/\gamma^2$. Thus, plugging (B.6) into (B.5), we have

$$
\begin{aligned}
|I_{22}| &\leq \frac{\beta}{2}\gamma\sigma_r^*\|\mathbf{H}\|_F^2 + \frac{2+\gamma}{2\beta}\|\widetilde{\mathbf{U}}\mathbf{H}_V^\top + \mathbf{H}_U\widetilde{\mathbf{V}}^\top\|_F^2 + \frac{\beta}{4}\|\mathbf{H}\|_F^4 \\
&\leq \frac{2+\gamma}{\beta}\|\mathbf{M} - \mathbf{M}^*\|_F^2 + \frac{\beta}{2}\gamma\sigma_r^*\|\mathbf{H}\|_F^2 + \left(\frac{2+\gamma}{4\beta} + \frac{\beta}{4}\right)\|\mathbf{H}\|_F^4,
\end{aligned} \tag{B.7}
$$

where $\mathbf{M} = \mathbf{U}\mathbf{V}^\top$. Similarly, according to Lemma B.3, we can upper bound $|I_{23}|$

$$
\begin{aligned}
|I_{23}| &\leq \frac{1}{p}|\langle \mathcal{P}_\Omega(\mathbf{X}_L\mathbf{H}_U\mathbf{H}_V^\top\mathbf{X}_R^\top), \mathbf{X}_L\mathbf{H}_U\mathbf{H}_V^\top\mathbf{X}_R^\top\rangle| + |\langle \mathbf{X}_L\mathbf{H}_U\mathbf{H}_V^\top\mathbf{X}_R^\top, \mathbf{X}_L\mathbf{H}_U\mathbf{H}_V^\top\mathbf{X}_R^\top\rangle| \\
&\leq \frac{1}{p}\|\mathcal{P}_\Omega(\mathbf{X}_L\mathbf{H}_U\mathbf{H}_V^\top\mathbf{X}_R^\top)\|_F^2 + \|\mathbf{H}_U\mathbf{H}_V^\top\|_F^2 \\
&\leq \frac{1}{2}\|\mathbf{H}\|_F^4 + \gamma\sigma_r^*\|\mathbf{H}\|_F^2.
\end{aligned} \tag{B.8}
$$

Therefore, combining (B.4),(B.7) and (B.8), we obtain the upper bound of $|I_2|$

$$|I_2| \leq |I_{21}| + 3|I_{22}| + 2|I_{23}|$$
$$\leq \left(2\gamma + \frac{3(2+\gamma)}{\beta}\right)\|\mathbf{M} - \mathbf{M}^*\|_F^2 + \left(\frac{\gamma}{2} + \frac{3(2+\gamma)}{4\beta} + \frac{3\beta}{4} + 1\right)\|\mathbf{H}\|_F^4 + \left(\frac{3\beta}{2} + 2\right)\gamma\sigma_r^*\|\mathbf{H}\|_F^2. \quad (\text{B.9})$$

Finally, set $\beta = 48$, then combining (B.3) and (B.9), we obtain

$$\langle \nabla \widetilde{f}_\Omega(\mathbf{Z}), \mathbf{H} \rangle \geq \left(\frac{1}{16} - \frac{3\gamma}{2}\right)\|\mathbf{ZZ}^\top - \mathbf{Z}^*\mathbf{Z}^{*\top}\|_F^2 + \frac{1}{4\sigma_1^*}\|\widetilde{\mathbf{Z}}^*\widetilde{\mathbf{Z}}^{*\top}\mathbf{Z}\|_F^2 + \left(\frac{\sigma_r^*}{4} - 75\gamma\sigma_r^*\right)\|\mathbf{H}\|_F^2 - 40\|\mathbf{H}\|_F^4,$$

where the inequality holds because $\|\mathbf{M} - \mathbf{M}^*\|_F^2 \leq \|\mathbf{ZZ}^\top - \mathbf{Z}^*\mathbf{Z}^{*\top}\|_F^2/2$. Thus, by choosing constant $\gamma$ to be sufficiently small, we complete the proof. □

## B.2 Proof of Lemma A.2

*Proof.* Recall the gradient of $\widetilde{f}_\Omega$ in (A.3), we have

$$\|\nabla \widetilde{f}_\Omega(\mathbf{Z})\|_F^2 \leq 2\underbrace{\left\|p^{-1}\mathbf{X}^\top \mathcal{P}_{\widetilde{\Omega}}(\mathbf{X}[\mathbf{ZZ}^\top - \mathbf{Z}^*\mathbf{Z}^{*\top}]\mathbf{X}^\top)\mathbf{XZ}\right\|_F^2}_{I_1} + \frac{1}{2}\underbrace{\left\|(\overline{\mathcal{P}}_{\text{diag}} - \overline{\mathcal{P}}_{\text{off}})\mathbf{ZZ}^\top\mathbf{Z}\right\|_F^2}_{I_2},$$

where the inequality holds because $\|\mathbf{A} + \mathbf{B}\|_F^2 \leq 2(\|\mathbf{A}\|_F^2 + \|\mathbf{B}\|_F^2)$. In the sequel, we will upper bound $I_1$ and $I_2$, respectively.

**Upper bound of $I_1$.** Recall that $\mathbf{Z} = [\mathbf{U}; \mathbf{V}]$, $\mathbf{R} = \text{argmin}_{\widetilde{\mathbf{R}} \in \mathbb{Q}_r} \|\mathbf{Z} - \mathbf{Z}^*\widetilde{\mathbf{R}}\|_F$, and $\mathbf{H} = \mathbf{Z} - \mathbf{Z}^*\mathbf{R}$. Denote $\mathbf{M} = \mathbf{U}\mathbf{V}^\top$, then based on the above notations, we have

$$I_1 = \|p^{-1}\mathbf{X}_L^\top \mathcal{P}_\Omega(\mathbf{X}_L[\mathbf{M} - \mathbf{M}^*]\mathbf{X}_R^\top)\mathbf{X}_R\mathbf{V}\|_F^2 + \|p^{-1}\mathbf{X}_R^\top(\mathcal{P}_\Omega(\mathbf{X}_L[\mathbf{M} - \mathbf{M}^*]\mathbf{X}_R^\top))^\top \mathbf{X}_L\mathbf{U}\|_F^2$$
$$\leq r \cdot \underbrace{\|p^{-1}\mathbf{X}_L^\top \mathcal{P}_\Omega(\mathbf{X}_L[\mathbf{M} - \mathbf{M}^*]\mathbf{X}_R^\top)\mathbf{X}_R\mathbf{V}\|_2^2}_{I_{11}} + r \cdot \underbrace{\|p^{-1}\mathbf{X}_R^\top(\mathcal{P}_\Omega(\mathbf{X}_L[\mathbf{M} - \mathbf{M}^*]\mathbf{X}_R^\top))^\top \mathbf{X}_L\mathbf{U}\|_2^2}_{I_{12}}, \quad (\text{B.10})$$

where the inequality holds because both $\mathbf{U}$ and $\mathbf{V}$ have rank at most $r$. Consider the term $I_{11}$ first, we observe

$$\frac{1}{p}\mathbf{X}_L^\top \mathcal{P}_\Omega(\mathbf{X}_L[\mathbf{M} - \mathbf{M}^*]\mathbf{X}_R^\top)\mathbf{X}_R\mathbf{V} - (\mathbf{M} - \mathbf{M}^*)\mathbf{V}$$
$$= \sum_{(i,j)=(1,1)}^{(d_1,d_2)} (p^{-1}\xi_{ij} - 1) \cdot [\mathbf{X}_L(\mathbf{M} - \mathbf{M}^*)\mathbf{X}_R^\top]_{ij} \cdot [\mathbf{X}_L]_{i,*}^\top \cdot [\mathbf{X}_R\mathbf{V}]_{j,*} := \sum_{(i,j)=(1,1)}^{(d_1,d_2)} \mathbf{A}_{ij},$$

where $\xi_{ij} = 1$, if $(i,j) \in \Omega$; $\xi_{ij} = 0$, otherwise. We are going to apply matrix bernstein inequality to the above summation. Due to sample splitting, the randomness only comes from $\Omega$, thus $\mathbf{A}_{ij}$'s are independent and $\mathbb{E}[\mathbf{A}_{ij}] = 0$. Denote $\widetilde{\mathbf{U}} = \mathbf{U}^*\mathbf{R}$ and $\widetilde{\mathbf{V}} = \mathbf{V}^*\mathbf{R}$, and let $\mathbf{H}_U \in \mathbb{R}^{n_1 \times r}$, $\mathbf{H}_V \in \mathbb{R}^{n_2 \times r}$ be the top $n_1$ and bottom $n_2$ rows of $\mathbf{H}$. Then we have $\mathbf{M} - \mathbf{M}^* = \widetilde{\mathbf{U}}\mathbf{H}_V^\top + \mathbf{H}_U\widetilde{\mathbf{V}}^\top + \mathbf{H}_U\mathbf{H}_V^\top$. For any $(i,j)$, we have the following upper bound of $\|\mathbf{A}_{ij}\|_2$

$$\|\mathbf{A}_{ij}\|_2 \leq \frac{1}{p}\left|[\mathbf{X}_L(\mathbf{M} - \mathbf{M}^*)\mathbf{X}_R^\top]_{ij}\right| \cdot \|[\mathbf{X}_L]_{i,*}^\top \cdot [\mathbf{X}_R\mathbf{V}]_{j,*}\|_2$$
$$\leq \frac{1}{p}\|\mathbf{X}_L(\widetilde{\mathbf{U}}\mathbf{H}_V^\top + \mathbf{H}_U\widetilde{\mathbf{V}}^\top + \mathbf{H}_U\mathbf{H}_V^\top)\mathbf{X}_R^\top\|_{\infty,\infty} \cdot \|\mathbf{X}_L\|_{2,\infty} \cdot \|\mathbf{X}_R\mathbf{V}\|_{2,\infty}$$
$$\leq \frac{10\mu_0\mu_1 rn\sigma_1^*}{pd_1d_2} \cdot \|\mathbf{H}\|_F,$$

where the last inequality is due to Assumptions 3.1, 3.2 and the fact that $\|\mathbf{X}_L\mathbf{H}_U\|_{2,\infty} \leq \|\mathbf{X}_L\mathbf{U}\|_{2,\infty} + \|\mathbf{X}_L\widetilde{\mathbf{U}}\|_{2,\infty} \leq 3\sqrt{\mu_0 r \sigma_1^*/d_1}$ and $\|\mathbf{X}_R\mathbf{V}\|_{2,\infty} \leq 2\sqrt{\mu_0 r \sigma_1^*/d_2}$. To apply matrix bernstein, it remains to bound $\|\sum_{i,j} \mathbb{E}[\mathbf{A}_{ij}\mathbf{A}_{ij}^\top]\|_2$ and $\|\sum_{i,j} \mathbb{E}[\mathbf{A}_{ij}^\top\mathbf{A}_{ij}]\|_2$. In particular, we have

$$\Big\|\sum_{(i,j)=(1,1)}^{(d_1,d_2)} \mathbb{E}[\mathbf{A}_{ij}\mathbf{A}_{ij}^\top]\Big\|_2 = \frac{1-p}{p} \Big\|\sum_{(i,j)=(1,1)}^{(d_1,d_2)} [\mathbf{X}_L(\mathbf{M}-\mathbf{M}^*)\mathbf{X}_R^\top]_{ij}^2 \cdot [\mathbf{X}_L]_{i,*}^\top \cdot \big\|[\mathbf{X}_R\mathbf{V}]_{j,*}\big\|_2^2 \cdot [\mathbf{X}_L]_{i,*}\Big\|_2$$

$$\leq \frac{1}{p} \Big\|\sum_{i=1}^{d_1} \mathbf{e}_i\mathbf{e}_i^\top \cdot \sum_{j=1}^{d_2} \Big([\mathbf{X}_L(\mathbf{M}-\mathbf{M}^*)\mathbf{X}_R^\top]_{ij}^2 \cdot \big\|[\mathbf{X}_R\mathbf{V}]_{j,*}\big\|_2^2\Big)\Big\|_2$$

$$\leq \frac{1}{p} \max_{i\in[d_1]} \sum_{j=1}^{d_2} \Big([\mathbf{X}_L(\mathbf{M}-\mathbf{M}^*)\mathbf{X}_R^\top]_{ij}^2 \cdot \|\mathbf{X}_R\mathbf{V}\|_{2,\infty}^2\Big)$$

$$\leq \frac{4\mu_0 r \sigma_1^*}{pd_2} \cdot \big\|\mathbf{X}_L(\mathbf{M}-\mathbf{M}^*)\mathbf{X}_R^\top\big\|_{2,\infty}^2 \leq \frac{c\mu_0\mu_1 r \sigma_1^* n}{pd_1d_2} \cdot \|\mathbf{M}-\mathbf{M}^*\|_F^2,$$

where the last inequality holds because $\|\mathbf{AB}\|_{2,\infty} \leq \|\mathbf{A}\|_{2,\infty} \cdot \|\mathbf{B}\|_F$. Similarly, we have

$$\Big\|\sum_{(i,j)=(1,1)}^{(d_1,d_2)} \mathbb{E}[\mathbf{A}_{ij}^\top\mathbf{A}_{ij}]\Big\|_2 \leq \frac{1}{p} \Big\|\sum_{(i,j)=(1,1)}^{(d_1,d_2)} [\mathbf{X}_L(\mathbf{M}-\mathbf{M}^*)\mathbf{X}_R^\top]_{ij}^2 \cdot [\mathbf{X}_R\mathbf{V}]_{j,*}^\top \cdot \big\|[\mathbf{X}_L]_{i,*}\big\|_2^2 \cdot [\mathbf{X}_R\mathbf{V}]_{j,*}\Big\|_2$$

$$\leq \frac{1}{p} \|\mathbf{X}_L\|_{2,\infty}^2 \cdot \|\mathbf{X}_R\mathbf{V}\|_{2,\infty}^2 \cdot \sum_{(i,j)=(1,1)}^{(d_1,d_2)} [\mathbf{X}_L(\mathbf{M}-\mathbf{M}^*)\mathbf{X}_R^\top]_{ij}^2$$

$$\leq \frac{4\mu_0\mu_1 r \sigma_1^* n}{pd_1d_2} \cdot \|\mathbf{M}-\mathbf{M}^*\|_F^2,$$

where the second inequality follows from the definition of spectral norm. Therefore, according to Lemma D.1, with probability at least $1-c/d$, we have

$$\Big\|\sum_{(i,j)=(1,1)}^{(d_1,d_2)} \mathbf{A}_{ij}\Big\|_2 \leq \gamma \sigma_r^{*1/2}\|\mathbf{M}-\mathbf{M}^*\|_F + \gamma^2 \sigma_r^*\|\mathbf{H}\|_F,$$

under condition that $|\Omega| \geq c'\mu_0\mu_1 rn\kappa/\gamma^2$, where $c, c'$ are both constants. Hence, by triangle's inequality, we obtain the upper bound of $I_{11}$

$$I_{11} \leq \big(\|(\mathbf{M}-\mathbf{M}^*)\mathbf{V}\|_2 + \gamma\sigma_r^{*1/2}\|\mathbf{M}-\mathbf{M}^*\|_F + \gamma^2\sigma_r^*\|\mathbf{H}\|_F\big)^2 \leq 8\sigma_1^*\|\mathbf{M}-\mathbf{M}^*\|_F^2 + \sigma_r^{*2}\|\mathbf{H}\|_F^2, \quad (B.11)$$

where the last inequality holds because $\|\mathbf{V}\|_2 \leq \|\mathbf{V}^*\|_2 + \|\mathbf{V}-\mathbf{V}^*\mathbf{R}\|_2 \leq 2\sqrt{\sigma_1^*}$ and $\gamma \in (0, 1/2)$. Similarly, we obtain the upper bound of $I_{12}$

$$I_{12} \leq \big(\|\mathbf{U}^\top(\mathbf{M}-\mathbf{M}^*)\|_2 + \gamma\sigma_r^{*1/2}\|\mathbf{M}-\mathbf{M}^*\|_F + \gamma^2\sigma_r^{*2}\|\mathbf{H}\|_F\big)^2 \leq 8\sigma_1^*\|\mathbf{M}-\mathbf{M}^*\|_F^2 + \sigma_r^{*2}\|\mathbf{H}\|_F^2. \quad (B.12)$$

Plugging (B.11) and (B.12) into (B.10), we have

$$I_1 \leq 2r \cdot (8\sigma_1^*\|\mathbf{M}-\mathbf{M}^*\|_F^2 + \sigma_r^{*2}\|\mathbf{H}\|_F^2). \quad (B.13)$$

**Upper bound of $I_2$.** As for $I_2$, we obtain

$$I_2 = \|\widetilde{\mathbf{Z}}^*\widetilde{\mathbf{Z}}^{*\top}\mathbf{Z} - (\overline{\mathcal{P}}_{\text{diag}} - \overline{\mathcal{P}}_{\text{off}})(\mathbf{Z}^*\mathbf{Z}^{*\top} - \mathbf{Z}\mathbf{Z}^\top)\mathbf{Z}\|_F^2$$

$$\leq 2\|\widetilde{\mathbf{Z}}^*\widetilde{\mathbf{Z}}^{*\top}\mathbf{Z}\|_F^2 + 2\|(\overline{\mathcal{P}}_{\text{diag}} - \overline{\mathcal{P}}_{\text{off}})(\mathbf{Z}^*\mathbf{Z}^{*\top} - \mathbf{Z}\mathbf{Z}^\top)\|_F^2 \cdot \|\mathbf{Z}\|_2^2$$

$$\leq 2\|\widetilde{\mathbf{Z}}^*\widetilde{\mathbf{Z}}^{*\top}\mathbf{Z}\|_F^2 + 8\sigma_1^*\|\mathbf{Z}\mathbf{Z}^\top - \mathbf{Z}^*\mathbf{Z}^{*\top}\|_F^2, \quad (B.14)$$

where the first inequality holds because $\|\mathbf{A} - \mathbf{B}\|_F^2 \leq 2(\|\mathbf{A}\|_F^2 + \|\mathbf{B}\|_F^2)$ and $\|\mathbf{AB}\|_F \leq \|\mathbf{A}\|_F \cdot \|\mathbf{B}\|_2$, and the second inequality holds because $\|\mathbf{Z}\|_2 \leq 2\sqrt{\sigma_1^*}$.

Finally, combining (B.13) and (B.14), we obtain

$$\|\nabla \widetilde{f}_\Omega(\mathbf{Z})\|_F^2 \leq 32r\sigma_1^*\|\mathbf{M} - \mathbf{M}^*\|_F^2 + 4r\sigma_r^{*2}\|\mathbf{H}\|_F^2 + \|\widetilde{\mathbf{Z}}^*\widetilde{\mathbf{Z}}^{*\top}\mathbf{Z}\|_F^2 + 4\sigma_1^*\|\mathbf{Z}\mathbf{Z}^\top - \mathbf{Z}^*\mathbf{Z}^{*\top}\|_F^2$$
$$\leq (16r+4)\sigma_1^*\|\mathbf{Z}\mathbf{Z}^\top - \mathbf{Z}^*\mathbf{Z}^{*\top}\|_F^2 + \|\widetilde{\mathbf{Z}}^*\widetilde{\mathbf{Z}}^{*\top}\mathbf{Z}\|_F^2 + 4r\sigma_r^{*2}\|\mathbf{H}\|_F^2,$$

where the second inequality holds because $\|\mathbf{M} - \mathbf{M}^*\|_F^2 \leq \|\mathbf{Z}\mathbf{Z}^\top - \mathbf{Z}^*\mathbf{Z}^{*\top}\|_F^2/2$, which completes the proof. $\square$

## B.3 Proof of Lemma A.3

*Proof of Lemma A.3.* Similar to the proof of Lemma A.1, we have

$$\langle \nabla \widetilde{f}_\Omega(\mathbf{Z}), \mathbf{H}\rangle = \underbrace{\frac{1}{2}\langle \nabla G(\mathbf{Z}), \mathbf{H}\rangle + \frac{1}{2}\langle \widetilde{\mathbf{Z}}^*\widetilde{\mathbf{Z}}^{*\top}\mathbf{Z}, \mathbf{H}\rangle}_{I_1} + \underbrace{\langle (p^{-1}\mathcal{P}_{\widetilde{\Omega}} - \mathcal{P}_{\text{off}})(\mathbf{X}[\mathbf{Z}\mathbf{Z}^\top - \mathbf{Z}^*\mathbf{Z}^{*\top}]\mathbf{X}^\top), \mathbf{X}\mathbf{H}(\mathbf{X}\mathbf{Z})^\top\rangle}_{I_2}.$$

According to Lemma B.1, we have

$$I_1 \geq \frac{\sigma_r^*}{4}\|\mathbf{H}\|_F^2 + \frac{1}{8}\|\mathbf{Z}\mathbf{Z}^\top - \mathbf{Z}^*\mathbf{Z}^{*\top}\|_F^2 + \frac{1}{4\sigma_1^*}\|\widetilde{\mathbf{Z}}^*\widetilde{\mathbf{Z}}^{*\top}\mathbf{Z}\|_F^2. \tag{B.15}$$

Recall the notations in the proof of Lemma A.1, we let $\mathbf{Z} = [\mathbf{U}; \mathbf{V}]$ and $\mathbf{M}^* = \mathbf{U}^*\mathbf{V}^{*\top}$, we denote $\mathbf{M} = \mathbf{U}\mathbf{V}^\top$, $\widetilde{\mathbf{U}} = \mathbf{U}^*\mathbf{R}$ and $\widetilde{\mathbf{V}} = \mathbf{V}^*\mathbf{R}$. Besides, let $\mathbf{H}_U \in \mathbb{R}^{n_1 \times r}$, $\mathbf{H}_V \in \mathbb{R}^{n_2 \times r}$ be the top $n_1$ and bottom $n_2$ rows of $\mathbf{H}$. Then we can reformulate $I_2$ as follows

$$I_2 = \underbrace{\langle (p^{-1}\mathcal{P}_\Omega - \mathcal{I})(\mathbf{X}_L[\widetilde{\mathbf{U}}\mathbf{H}_V^\top + \mathbf{H}_U\widetilde{\mathbf{V}}^\top]\mathbf{X}_R^\top), \mathbf{X}_L(\widetilde{\mathbf{U}}\mathbf{H}_V^\top + \mathbf{H}_U\widetilde{\mathbf{V}}^\top)\mathbf{X}_R^\top\rangle}_{I_{21}}$$
$$+ 3\underbrace{\langle (p^{-1}\mathcal{P}_\Omega - \mathcal{I})(\mathbf{X}_L[\widetilde{\mathbf{U}}\mathbf{H}_V^\top + \mathbf{H}_U\widetilde{\mathbf{V}}^\top]\mathbf{X}_R^\top), \mathbf{X}_L\mathbf{H}_U\mathbf{H}_V^\top\mathbf{X}_R^\top\rangle}_{I_{22}}$$
$$+ 2\underbrace{\langle (p^{-1}\mathcal{P}_\Omega - \mathcal{I})(\mathbf{X}_L\mathbf{H}_U\mathbf{H}_V^\top\mathbf{X}_R^\top), \mathbf{X}_L\mathbf{H}_U\mathbf{H}_V^\top\mathbf{X}_R^\top\rangle}_{I_{23}}.$$

Note that $\widetilde{\mathbf{U}}\mathbf{H}_V^\top + \mathbf{H}_U\widetilde{\mathbf{V}}^\top$ falls into the subspace $\mathcal{T}$ defined in Lemma B.2. Thus according to Lemma B.2, we can still obtain the same upper bound of $I_{21}$

$$|I_{21}| \leq \gamma\|\widetilde{\mathbf{U}}\mathbf{H}_V^\top + \mathbf{H}_U\widetilde{\mathbf{V}}^\top\|_F^2 \leq 2\gamma\|\mathbf{M} - \mathbf{M}^*\|_F^2 + \frac{\gamma}{2}\|\mathbf{H}\|_F^4$$

holds with probability at least $1 - c_2/d$, provided that $|\Omega| \geq c_1\mu_0\mu_1 rn \log d/\gamma^2$. As for the term $I_{22}$, similarly we have

$$|I_{22}| \leq \frac{1}{p}\Big|\langle \mathcal{P}_\Omega(\mathbf{X}_L[\widetilde{\mathbf{U}}\mathbf{H}_V^\top + \mathbf{H}_U\widetilde{\mathbf{V}}^\top]\mathbf{X}_R^\top), \mathbf{X}_L\mathbf{H}_U\mathbf{H}_V^\top\mathbf{X}_R^\top\rangle\Big| + \Big|\langle \mathbf{X}_L(\widetilde{\mathbf{U}}\mathbf{H}_V^\top + \mathbf{H}_U\widetilde{\mathbf{V}}^\top)\mathbf{X}_R^\top, \mathbf{X}_L\mathbf{H}_U\mathbf{H}_V^\top\mathbf{X}_R^\top\rangle\Big|$$
$$\leq \frac{\beta}{2} \cdot p^{-1}\big\|\mathcal{P}_\Omega(\mathbf{X}_L\mathbf{H}_U\mathbf{H}_V^\top\mathbf{X}_R^\top)\big\|_F^2 + \frac{2+\gamma}{2\beta}\|\widetilde{\mathbf{U}}\mathbf{H}_V^\top + \mathbf{H}_U\widetilde{\mathbf{V}}^\top\|_F^2 + \frac{\beta}{8}\|\mathbf{H}\|_F^4,$$

where constant $\beta > 0$ will be specified later. Here, the second inequality follows from the Young's Inequality and Lemma B.2. According to bernstein-type inequality for Bernoulli random variables, we further obtain

$$\frac{1}{p}\big\|\mathcal{P}_\Omega(\mathbf{X}_L\mathbf{H}_U\mathbf{H}_V^\top\mathbf{X}_R^\top)\big\|_F^2 \leq \frac{1}{p}|\Omega| \cdot \|\mathbf{X}_L\mathbf{H}_U\mathbf{H}_V^\top\mathbf{X}_R^\top\|_{\infty,\infty}^2 \leq \frac{3}{2}\mu_1^2 n_1 n_2 \cdot \|\mathbf{H}_U\mathbf{H}_V^\top\|_F^2 \leq c_0^2\sigma_r^*\|\mathbf{H}\|_F^2$$

holds with probability at least $1 - c_1/d$, where the second inequality follows from Lemma D.2 and the incoherence Assumptions 3.2, and the last inequality holds because $\|\mathbf{H}\|_F^2 = D^2(\mathbf{Z}, \mathbf{Z}^*) \leq c_0^2 \sigma_r^*/(\mu_1^2 n^2)$. Therefore, we obtain the upper bound of $I_{22}$

$$|I_{22}| \leq \frac{2+\gamma}{2\beta} \|\widetilde{\mathbf{U}}\mathbf{H}_V^\top + \mathbf{H}_U \widetilde{\mathbf{V}}^\top\|_F^2 + \frac{\beta}{2} \cdot c_0^2 \sigma_r^* \|\mathbf{H}\|_F^2 + \frac{\beta}{8} \|\mathbf{H}\|_F^4$$
$$\leq \frac{2+\gamma}{\beta} \|\mathbf{M} - \mathbf{M}^*\|_F^2 + \frac{\beta}{2} \cdot c_0^2 \sigma_r^* \|\mathbf{H}\|_F^2 + \left(\frac{2+\gamma}{4\beta} + \frac{\beta}{8}\right) \|\mathbf{H}\|_F^4.$$

Similarly, we can upper bound $I_{23}$

$$|I_{23}| \leq \frac{1}{p} |\langle \mathcal{P}_\Omega(\mathbf{X}_L \mathbf{H}_U \mathbf{H}_V^\top \mathbf{X}_R^\top), \mathbf{X}_L \mathbf{H}_U \mathbf{H}_V^\top \mathbf{X}_R^\top \rangle| + |\langle \mathbf{X}_L \mathbf{H}_U \mathbf{H}_V^\top \mathbf{X}_R^\top, \mathbf{X}_L \mathbf{H}_U \mathbf{H}_V^\top \mathbf{X}_R^\top \rangle|$$
$$= \frac{1}{p} \|\mathcal{P}_\Omega(\mathbf{X}_L \mathbf{H}_U \mathbf{H}_V^\top \mathbf{X}_R^\top)\|_F^2 + \|\mathbf{H}_U \mathbf{H}_V^\top\|_F^2 \leq c_0^2 \sigma_r^* \|\mathbf{H}\|_F^2 + \frac{1}{4} \|\mathbf{H}\|_F^4.$$

Hence, we obtain

$$|I_2| \leq |I_{21}| + 3|I_{22}| + 2|I_{23}|$$
$$\leq \left(2\gamma + \frac{3(2+\gamma)}{\beta}\right) \cdot \|\mathbf{M} - \mathbf{M}^*\|_F^2 + \left(2 + \frac{3\beta}{2}\right) \cdot c_0^2 \sigma_r^* \|\mathbf{H}\|_F^2 + \left(\frac{3(2+\gamma)}{4\beta} + \frac{3\beta}{8} + \frac{1}{2}\right) \cdot \|\mathbf{H}\|_F^4. \quad \text{(B.16)}$$

Set $\beta = 48$, and let constants $c_0, \gamma$ be sufficiently small. Combining (B.15) and (B.16), we complete the proof. □

### B.4 Proof of Lemma A.4

In order to prove Lemma A.4, we need to make use of the following lemma.

**Lemma B.4.** Let $\mathbf{X} \in \mathbb{R}^{d_1 \times n_1}$, $\mathbf{Y} \in \mathbb{R}^{d_2 \times n_2}$ be the feature matrices, which are orthonormal and self-incoherent with parameter $\mu_1$, and $\Omega \subseteq [d_1] \times [d_2]$ be an index set followed Bernoulli Model (3.1) with $p = |\Omega|/(d_1 d_2)$. For any $\gamma \in (0,1)$, there exist constants $c_1$ and $c_2$ such that, under condition $|\Omega| \geq c_1 \mu_1 n \log d / \gamma^2$, for all $\mathbf{Z} \in \mathbb{R}^{n_1 \times n_2}$

$$\|p^{-1} \mathbf{X}^\top \mathcal{P}_\Omega (\mathbf{X} \mathbf{Z} \mathbf{Y}^\top) \mathbf{Y}\|_F \leq (1 + \mu_1 n \gamma) \|\mathbf{Z}\|_F,$$

holds with probability at least $1 - c_2/d$.

Now we are ready to prove Lemma A.4.

*Proof of Lemma A.4.* According to the gradient of $\widetilde{f}_\Omega$ in (A.3), we have

$$\|\nabla \widetilde{f}_\Omega(\mathbf{Z})\|_F^2 \leq 2 \underbrace{\left\| p^{-1} \mathbf{X}^\top \mathcal{P}_{\widetilde{\Omega}}\left(\mathbf{X}[\mathbf{Z}\mathbf{Z}^\top - \mathbf{Z}^* \mathbf{Z}^{*\top}] \mathbf{X}^\top\right) \mathbf{X}\mathbf{Z} \right\|_F^2}_{I_1} + \frac{1}{2} \underbrace{\left\| (\overline{\mathcal{P}}_{\text{diag}} - \overline{\mathcal{P}}_{\text{off}}) \mathbf{Z}\mathbf{Z}^\top \mathbf{Z} \right\|_F^2}_{I_2}.$$

In the following discussions, we will upper bound $I_1$ and $I_2$ respectively. As for $I_1$, similar to the proof of Lemma A.2, we have

$$I_1 = \underbrace{\left\| p^{-1} \mathbf{X}_L^\top \mathcal{P}_\Omega(\mathbf{X}_L [\mathbf{M} - \mathbf{M}^*] \mathbf{X}_R^\top) \mathbf{X}_R \mathbf{V} \right\|_F^2}_{I_{11}} + \underbrace{\left\| p^{-1} \mathbf{X}_R^\top \left(\mathcal{P}_\Omega(\mathbf{X}_L [\mathbf{M} - \mathbf{M}^*] \mathbf{X}_R^\top)\right)^\top \mathbf{X}_L \mathbf{U} \right\|_F^2}_{I_{12}}. \quad \text{(B.17)}$$

Consider the first term $I_{11}$. Note that $\mathbf{M} - \mathbf{M}^* = \mathbf{H}_U \widetilde{\mathbf{V}}^\top + \widetilde{\mathbf{U}} \mathbf{H}_V^\top + \mathbf{H}_U \mathbf{H}_V^\top$. Recall the definition of $\mathcal{P}_\mathcal{T}$ in Lemma B.2, we have $\mathcal{P}_\mathcal{T}(\mathbf{M}) + \mathcal{P}_{\mathcal{T}^\perp}(\mathbf{M}) = \mathbf{M}$, for any $\mathbf{M} \in \mathbb{R}^{n_1 \times n_2}$, where $\mathcal{P}_{\mathcal{T}^\perp}(\mathbf{M}) = (\mathbf{I} - \overline{\mathbf{U}}^* \overline{\mathbf{U}}^{*\top}) \mathbf{M} (\mathbf{I} -$

$\overline{\mathbf{V}}^*\overline{\mathbf{V}}^{*\top}$). Thus we obtain

$$I_{11} \leq 2\|p^{-1}(\mathbf{X}_L^\top \mathcal{P}_\Omega(\mathbf{X}_L[\widetilde{\mathbf{U}}\mathbf{H}_V^\top + \mathbf{H}_U\widetilde{\mathbf{V}}^\top]\mathbf{X}_R^\top)\mathbf{X}_R)\mathbf{V}\|_F^2 + 2\|p^{-1}\mathbf{X}_L^\top \mathcal{P}_\Omega(\mathbf{X}_L\mathbf{H}_U\mathbf{H}_V^\top\mathbf{X}_R^\top)\mathbf{X}_R\mathbf{V}\|_F^2$$

$$\leq 4\underbrace{\|p^{-1}\mathcal{P}_\mathcal{T}(\mathbf{X}_L^\top \mathcal{P}_\Omega(\mathbf{X}_L[\widetilde{\mathbf{U}}\mathbf{H}_V^\top + \mathbf{H}_U\widetilde{\mathbf{V}}^\top]\mathbf{X}_R^\top)\mathbf{X}_R)\mathbf{V}\|_F^2}_{J_1}$$

$$+ 4\underbrace{\|p^{-1}\mathcal{P}_{\mathcal{T}^\perp}(\mathbf{X}_L^\top \mathcal{P}_\Omega(\mathbf{X}_L[\widetilde{\mathbf{U}}\mathbf{H}_V^\top + \mathbf{H}_U\widetilde{\mathbf{V}}^\top]\mathbf{X}_R^\top)\mathbf{X}_R)\mathbf{H}_V\|_F^2}_{J_2} + 2\underbrace{\|p^{-1}\mathbf{X}_L^\top \mathcal{P}_\Omega(\mathbf{X}_L\mathbf{H}_U\mathbf{H}_V^\top\mathbf{X}_R^\top)\mathbf{X}_R\mathbf{V}\|_F^2}_{J_3},$$

where the inequality follows from $\|\mathbf{A}+\mathbf{B}\|_F^2 \leq 2\|\mathbf{A}\|_F^2 + 2\|\mathbf{B}\|_F^2$, and the equality is due to the definition of $\mathcal{P}_{\mathcal{T}^\perp}$ and the fact that $(\mathbf{I} - \overline{\mathbf{V}}^*\overline{\mathbf{V}}^{*\top})\widetilde{\mathbf{V}} = \mathbf{0}$. In the sequel, we will upper bound the terms $J_1$, $J_2$ and $J_3$ respectively. Note that $\widetilde{\mathbf{U}}\mathbf{H}_V^\top + \mathbf{H}_U\widetilde{\mathbf{V}}^\top \in \mathcal{T}$, according to Lemma B.2, with probability at least $1 - c'/d$, we have

$$J_1 \leq (1+\gamma)^2\|\widetilde{\mathbf{U}}\mathbf{H}_V^\top + \mathbf{H}_U\widetilde{\mathbf{V}}^\top\|_F^2 \cdot \|\mathbf{V}\|_2^2 \leq 4(1+\gamma)^2\sigma_1^* \cdot \|\widetilde{\mathbf{U}}\mathbf{H}_V^\top + \mathbf{H}_U\widetilde{\mathbf{V}}^\top\|_F^2, \tag{B.18}$$

provided that $|\Omega| \geq \mu_0\mu_1 rn \log d/\gamma^2$, where the first inequality holds because $\|\mathbf{AB}\|_F \leq \|\mathbf{A}\|_F \cdot \|\mathbf{B}\|_2$, and the second inequality follows from the fact that $\|\mathbf{V}\|_2 \leq \|\mathbf{V}^*\|_2 + \|\mathbf{H}_V\|_2 \leq 2\sqrt{\sigma_1^*}$. Consider the second term $J_2$, we have

$$J_2 \leq \|p^{-1}\mathcal{P}_{\mathcal{T}^\perp}(\mathbf{X}_L^\top \mathcal{P}_\Omega(\mathbf{X}_L[\widetilde{\mathbf{U}}\mathbf{H}_V^\top + \mathbf{H}_U\widetilde{\mathbf{V}}^\top]\mathbf{X}_R^\top)\mathbf{X}_R)\|_F^2 \cdot \|\mathbf{H}_V\|_F^2$$

$$\leq \|p^{-1}\mathbf{X}_L^\top \mathcal{P}_\Omega(\mathbf{X}_L[\widetilde{\mathbf{U}}\mathbf{H}_V^\top + \mathbf{H}_U\widetilde{\mathbf{V}}^\top]\mathbf{X}_R^\top)\mathbf{X}_R\|_F^2 \cdot \|\mathbf{H}_V\|_F^2$$

$$\leq (1+\mu_1 n\gamma)^2 \cdot \|\widetilde{\mathbf{U}}\mathbf{H}_V^\top + \mathbf{H}_U\widetilde{\mathbf{V}}^\top\|_F^2 \cdot \|\mathbf{H}_V\|_F^2 \leq 2\gamma^2\sigma_r^*\|\widetilde{\mathbf{U}}\mathbf{H}_V^\top + \mathbf{H}_U\widetilde{\mathbf{V}}^\top\|_F^2,$$

where the first inequality holds because $\|\mathbf{AB}\|_F \leq \|\mathbf{A}\|_F \cdot \|\mathbf{B}\|_F$, the second inequality is due to the non-expansiveness of $\mathcal{P}_{\mathcal{T}^\perp}$, the third inequality follows from Lemma B.4, and the last inequality is due to $\|\mathbf{H}\|_F \leq c_0\sqrt{\sigma_r^*}/(\mu_1 n)$. According to Lemma B.4, we can upper bound the last term $J_3$ as follows

$$J_3 \leq \|p^{-1}\mathbf{X}_L^\top \mathcal{P}_\Omega(\mathbf{X}_L\mathbf{H}_U\mathbf{H}_V^\top\mathbf{X}_R^\top)\mathbf{X}_R\|_F^2 \cdot \|\mathbf{V}\|_2^2$$

$$\leq (1+\mu_1 n\gamma)^2 \cdot \|\mathbf{H}_U\mathbf{H}_V^\top\|_F^2 \cdot \|\mathbf{V}\|_2^2 \leq 2\gamma^2\sigma_1^*\sigma_r^*\|\mathbf{H}\|_F^2,$$

where the second inequality follows from Lemma B.4, and the last inequality is due to $\|\mathbf{H}_U\mathbf{H}_V^\top\|_F \leq \|\mathbf{H}\|_F^2/2$ and $\|\mathbf{H}\|_F \leq \alpha\sqrt{\sigma_r^*}/(\mu_1 n)$. Therefore, we obtain the upper bound of $I_{11}$

$$I_{11} \leq 4J_1 + 4J_2 + 2J_3 \leq 20\sigma_1^*\|\widetilde{\mathbf{U}}\mathbf{H}_V^\top + \mathbf{H}_U\widetilde{\mathbf{V}}^\top\|_F^2 + \sigma_1^*\sigma_r^*\|\mathbf{H}\|_F^2$$

$$\leq 40\sigma_1^*\|\mathbf{M} - \mathbf{M}^*\|_F^2 + 10\sigma_1^*\|\mathbf{H}\|_F^4 + \sigma_1^*\sigma_r^*\|\mathbf{H}\|_F^2,$$

where we set $\gamma$ to be small enough in the first inequality, and the last inequality follows from $\|\mathbf{A}+\mathbf{B}\|_F^2 \leq 2\|\mathbf{A}\|_F^2 + 2\|\mathbf{B}\|_F^2$. By symmetry, we can use the same techniques to bound the term $I_{12}$, which will yields the same upper bound and implies

$$I_1 \leq 80\sigma_1^*\|\mathbf{M} - \mathbf{M}^*\|_F^2 + 20\sigma_1^*\|\mathbf{H}\|_F^4 + 2\sigma_1^*\sigma_r^*\|\mathbf{H}\|_F^2.$$

The upper bound of the remaining term $I_2$ is as follows

$$I_2 \leq 2\|\widetilde{\mathbf{Z}}^*\widetilde{\mathbf{Z}}^{*\top}\mathbf{Z}\|_F^2 + 8\sigma_1^*\|\mathbf{Z}\mathbf{Z}^\top - \mathbf{Z}^*\mathbf{Z}^{*\top}\|_F^2,$$

where the detailed proof can be found in the proof of Lemma A.2. Hence, we obtain

$$\|\nabla\widetilde{f}_\Omega(\mathbf{Z})\|_F^2 \leq 2I_1 + \frac{1}{2}I_2 \leq 84\sigma_1^*\|\mathbf{Z}\mathbf{Z}^\top - \mathbf{Z}^*\mathbf{Z}^{*\top}\|_F^2 + \|\widetilde{\mathbf{Z}}^*\widetilde{\mathbf{Z}}^{*\top}\mathbf{Z}\|_F^2 + 4\sigma_1^*\sigma_r^*\|\mathbf{H}\|_F^2 + 40\sigma_1^*\|\mathbf{H}\|_F^4,$$

which completes the proof. $\square$

# C Proof of Technical Lemmas

In this section, we prove the technical lemmas used in Section B.

## C.1 Proof of Lemma B.1

*Proof.* We begin our proof with some properties regarding the optimal solution. Let $\mathbf{H} = \mathbf{Z} - \mathbf{Z}^*\mathbf{R}$ and $\mathbf{A}\mathbf{\Sigma}\mathbf{B}^\top$ be the SVD of $\mathbf{Z}^{*\top}\mathbf{Z}$, then we have $\mathbf{R} = \mathbf{A}\mathbf{B}^\top$. Thus,

$$\mathbf{Z}^\top\mathbf{Z}^*\mathbf{R} = \mathbf{B}\mathbf{\Sigma}\mathbf{B}^\top = (\mathbf{Z}^*\mathbf{R})^\top\mathbf{Z},$$

which implies that $\mathbf{Z}^\top\mathbf{Z}^*\mathbf{R}$ is symmetric and positive definite. Moreover, we have

$$\mathbf{H}^\top\mathbf{Z}^*\mathbf{R} = \mathbf{Z}^\top\mathbf{Z}^*\mathbf{R} - \mathbf{R}^\top\mathbf{Z}^{*\top}\mathbf{Z}^*\mathbf{R} = \mathbf{Z}^\top\mathbf{Z}^*\mathbf{R} - (\mathbf{Z}^*\mathbf{R})^\top\mathbf{Z}^*\mathbf{R},$$

which implies $\mathbf{H}^\top\mathbf{Z}^*\mathbf{R}$ is also symmetric. Without loss of generality, we assume $\mathbf{R} = \mathbf{I}$, then $\mathbf{Z}^\top\mathbf{Z}^*$ is positive definite, and $\mathbf{H}^\top\mathbf{Z}^*$ is symmetric. Thus, to prove Lemma B.1, it is sufficient to prove

$$\langle(\mathbf{Z}^*\mathbf{H}^\top + \mathbf{H}\mathbf{Z}^{*\top} + \mathbf{H}\mathbf{H}^\top)(\mathbf{Z}^* + \mathbf{H}), \mathbf{H}\rangle \geq \frac{\sigma_r^2(\mathbf{Z}^*)}{4}\|\mathbf{H}\|_F^2 + \frac{1}{4}\|\mathbf{Z}^*\mathbf{H}^\top + \mathbf{H}\mathbf{Z}^{*\top} + \mathbf{H}\mathbf{H}^*\|_F^2,$$

which is equivalent to

$$0 \leq \operatorname{tr}\Big((\mathbf{H}^\top\mathbf{H})^2 + 3\mathbf{H}^\top\mathbf{H}\mathbf{H}^\top\mathbf{Z}^* + (\mathbf{H}^\top\mathbf{Z}^*)^2 + \mathbf{H}^\top\mathbf{H}\mathbf{Z}^{*\top}\mathbf{Z}^*$$
$$- \frac{1}{4}[(\mathbf{H}^\top\mathbf{H})^2 + 4\mathbf{H}^\top\mathbf{H}\mathbf{H}^\top\mathbf{Z}^* + 2(\mathbf{H}^\top\mathbf{Z}^*)^2 + 2\mathbf{H}^\top\mathbf{H}\mathbf{Z}^{*\top}\mathbf{Z}^*] - \frac{\sigma_r^2(\mathbf{Z}^*)}{4}\mathbf{H}^\top\mathbf{H}\Big).$$

Combining terms, we have

$$0 \leq \operatorname{tr}\Big(\frac{3}{4}(\mathbf{H}^\top\mathbf{H})^2 + 2\mathbf{H}^\top\mathbf{H}\mathbf{H}^\top\mathbf{Z}^* + \frac{1}{2}(\mathbf{H}^\top\mathbf{Z}^*)^2 + \frac{1}{2}\mathbf{H}^\top\mathbf{H}\mathbf{Z}^{*\top}\mathbf{Z}^* - \frac{\sigma_r^2(\mathbf{Z}^*)}{4}\mathbf{H}^\top\mathbf{H}\Big),$$

which is further equivalent to

$$0 \leq \operatorname{tr}\Big(\frac{1}{2}(\mathbf{H}^\top\mathbf{Z}^* + 2\mathbf{H}^\top\mathbf{H})^2 - \frac{5}{4}(\mathbf{H}^\top\mathbf{H})^2 + \frac{1}{2}\mathbf{H}^\top\mathbf{H}\mathbf{Z}^{*\top}\mathbf{Z}^* - \frac{\sigma_r^2(\mathbf{Z}^*)}{4}\mathbf{H}^\top\mathbf{H}\Big). \quad \text{(C.1)}$$

Note that $\operatorname{tr}\big((\mathbf{H}^\top\mathbf{H})^2\big) = \|\mathbf{H}^\top\mathbf{H}\|_F^2 \leq \|\mathbf{H}\|_F^2 \cdot \|\mathbf{H}\|_2^2$, and $\operatorname{tr}(\mathbf{H}^\top\mathbf{H}\mathbf{Z}^{*\top}\mathbf{Z}^*) \geq \sigma_r^2(\mathbf{Z}^*) \cdot \|\mathbf{H}\|_F^2$. Therefore, in order to prove (C.1), it is sufficient to require that

$$\|\mathbf{H}\|_2^2 \leq \frac{\sigma_r^2(\mathbf{Z}^*)}{5},$$

which completes the proof. □

## C.2 Proof of Lemma B.2

*Proof.* For any $i \in [d_1]$, denote $\mathbf{X}_{i,*} \in \mathbb{R}^{1 \times n_1}$ as the $i$-th row of $\mathbf{X}$. Similarly, for any $j \in [d_2]$, denote $\mathbf{Y}_{j,*} \in \mathbb{R}^{1 \times n_2}$ as the $j$-th row of $\mathbf{Y}$. Thus, for any $\mathbf{Z} \in \mathbb{R}^{n_1 \times n_2}$, we have

$$\begin{aligned}
\mathbf{X}^\top \mathcal{P}_\Omega\big(\mathbf{X}\mathcal{P}_\mathcal{T}(\mathbf{Z})\mathbf{Y}^\top\big)\mathbf{Y} &= \sum_{(i,j)\in\Omega} \langle\mathbf{X}\mathcal{P}_\mathcal{T}(\mathbf{Z})\mathbf{Y}^\top, \mathbf{e}_i\mathbf{e}_j^\top\rangle \mathbf{X}^\top\mathbf{e}_i\mathbf{e}_j^\top\mathbf{Y} \\
&= \sum_{(i,j)\in\Omega} \langle\mathcal{P}_\mathcal{T}(\mathbf{Z}), \mathbf{X}_{i,*}^\top\mathbf{Y}_{j,*}\rangle \mathbf{X}_{i,*}^\top\mathbf{Y}_{j,*} \\
&= \sum_{(i,j)\in\Omega} \langle\mathbf{Z}, \mathcal{P}_\mathcal{T}(\mathbf{X}_{i,*}^\top\mathbf{Y}_{j,*})\rangle \mathbf{X}_{i,*}^\top\mathbf{Y}_{j,*}, \quad \text{(C.2)}
\end{aligned}$$

where the last equality holds because $\langle \mathcal{P}_\mathcal{T}(\mathbf{A}), \mathbf{B} \rangle = \langle \mathbf{A}, \mathcal{P}_\mathcal{T}(\mathbf{B}) \rangle$. Besides, for any $(i,j) \in [d_1] \times [d_2]$, let $\xi_{ij} = 1$, if $(i,j) \in \Omega$, and zero otherwise. Note that both $\mathbf{X}$ and $\mathbf{Y}$ are orthonormal, thus according to (C.2), we have

$$\mathcal{P}_\mathcal{T}(\mathbf{Z}) - p^{-1}\mathcal{P}_\mathcal{T}\Big(\mathbf{X}^\top \mathcal{P}_\Omega\big(\mathbf{X}\mathcal{P}_\mathcal{T}(\mathbf{Z})\mathbf{Y}^\top\big)\mathbf{Y}\Big) = \mathcal{P}_\mathcal{T}\Big(\mathbf{X}^\top (\mathcal{I} - p^{-1}\mathcal{P}_\Omega)(\mathbf{X}\mathcal{P}_\mathcal{T}(\mathbf{Z})\mathbf{Y}^\top)\mathbf{Y}\Big)$$
$$= \sum_{\substack{(i,j)\in \\ [d_1]\times[d_2]}} (1 - p^{-1}\xi_{ij})\langle \mathbf{Z}, \mathcal{P}_\mathcal{T}(\mathbf{X}_{i,*}^\top \mathbf{Y}_{j,*})\rangle \mathcal{P}_\mathcal{T}(\mathbf{X}_{i,*}^\top \mathbf{Y}_{j,*}),$$

where $\mathcal{I}: \mathbb{R}^{d_1 \times d_2} \to \mathbb{R}^{d_1 \times d_2}$ is the identity mapping. For any $(i,j) \in [d_1] \times [d_2]$, define linear operator $\mathcal{S}_{ij} : \mathbb{R}^{n_1 \times n_2} \to \mathbb{R}^{n_1 \times n_2}$, such that

$$\mathcal{S}_{ij}(\mathbf{Z}) = (1 - p^{-1}\xi_{ij})\langle \mathbf{Z}, \mathcal{P}_\mathcal{T}(\mathbf{X}_{i,*}^\top \mathbf{Y}_{j,*})\rangle \mathcal{P}_\mathcal{T}(\mathbf{X}_{i,*}^\top \mathbf{Y}_{j,*}).$$

Define $\mathbf{S}_{ij} \in \mathbb{R}^{d_1 d_2 \times d_1 d_2}$ as the corresponding matrix to the linear operator $\mathcal{S}_{ij}$ such that

$$\mathbf{S}_{ij} = (1 - p^{-1}\xi_{ij})\text{vec}\big(\mathcal{P}_\mathcal{T}(\mathbf{X}_{i,*}^\top \mathbf{Y}_{j,*})\big)\text{vec}\big(\mathcal{P}_\mathcal{T}(\mathbf{X}_{i,*}^\top \mathbf{Y}_{j,*})\big)^\top,$$

then we can easily show that $\|\sum_{i,j} \mathcal{S}_{ij}(\mathbf{Z})\|_F^2 = \|\sum_{i,j} \mathbf{S}_{ij} \cdot \text{vec}(\mathbf{Z})\|_F^2$, for any $\mathbf{Z} \in \mathbb{R}^{d_1 \times d_2}$. Next, we are going to apply matrix bernstein to the summation $\sum_{i,j} \mathbf{S}_{ij}$. We note that $\mathbb{E}[\mathbf{S}_{ij}] = \mathbf{0}$, and $\mathbf{S}_{ij}$ is symmetric, for any $(i,j) \in [d_1] \times [d_2]$. Since $\mathbf{X}\mathbf{M}^*\mathbf{Y}^\top$ is $\mu_0$-incoherent and $\mathbf{M}^* = \overline{\mathbf{U}}^*\boldsymbol{\Sigma}\overline{\mathbf{V}}^{*\top}$, we have $\|\mathbf{X}\overline{\mathbf{U}}^*\|_{2,\infty} \leq \sqrt{\mu_0 r/d_1}$, $\|\mathbf{Y}\overline{\mathbf{V}}^*\|_{2,\infty} \leq \sqrt{\mu_0 r/d_2}$. According to the definition of $\mathcal{P}_\mathcal{T}$ in (B.1), for any $(i,j) \in [d_1] \times [d_2]$, we obtain

$$\|\mathcal{P}_\mathcal{T}(\mathbf{X}_{i,*}^\top \mathbf{Y}_{j,*})\|_F^2 = \langle \overline{\mathbf{U}}^*\overline{\mathbf{U}}^{*\top}\mathbf{X}_{i,*}^\top\mathbf{Y}_{j,*} + \mathbf{X}_{i,*}^\top\mathbf{Y}_{j,*}\overline{\mathbf{V}}^*\overline{\mathbf{V}}^{*\top} - \overline{\mathbf{U}}^*\overline{\mathbf{U}}^{*\top}\mathbf{X}_{i,*}^\top\mathbf{Y}_{j,*}\overline{\mathbf{V}}^*\overline{\mathbf{V}}^{*\top}, \mathbf{X}_{i,*}^\top\mathbf{Y}_{j,*}\rangle$$
$$= \|\overline{\mathbf{U}}^{*\top}\mathbf{X}_{i,*}^\top\mathbf{Y}_{j,*}\|_F^2 + \|\overline{\mathbf{V}}^{*\top}\mathbf{Y}_{j,*}^\top\mathbf{X}_{i,*}\|_F^2 - \|\overline{\mathbf{U}}^{*\top}\mathbf{X}_{i,*}^\top\mathbf{Y}_{j,*}\overline{\mathbf{V}}^*\|_F^2$$
$$\leq \|\mathbf{X}_{i,*}\overline{\mathbf{U}}^*\|_2^2 \cdot \|\mathbf{Y}_{j,*}\|_2^2 + \|\mathbf{Y}_{j,*}\overline{\mathbf{V}}^*\|_2 \cdot \|\mathbf{X}_{i,*}\|_2^2$$
$$\leq \|\mathbf{X}\overline{\mathbf{U}}^*\|_{2,\infty}^2 \cdot \|\mathbf{Y}\|_{2,\infty}^2 + \|\mathbf{Y}\overline{\mathbf{V}}^*\|_{2,\infty} \cdot \|\mathbf{X}\|_{2,\infty}^2 \leq \frac{\mu_0 \mu_1 r(n_1 + n_2)}{d_1 d_2}, \quad (\text{C.3})$$

where the first equality holds because $\|\mathcal{P}_\mathcal{T}(\mathbf{A})\|_F^2 = \langle \mathcal{P}_\mathcal{T}(\mathbf{A}), \mathbf{A} \rangle$, the first inequality holds because for any vectors $\mathbf{x}, \mathbf{y} \in \mathbb{R}^n$, $\|\mathbf{x}\mathbf{y}^\top\|_F \leq \|\mathbf{x}\|_2 \cdot \|\mathbf{y}\|_2$, and the last inequality holds because both $\mathbf{X}, \mathbf{Y}$ are $\mu_1$ self-incoherent. To apply matrix bernstein inequality, we need to bound $\|\mathbf{S}_{ij}\|_2$ and $\|\sum_{i,j} \mathbb{E}(\mathbf{S}_{ij}^2)\|_2$, respectively. To begin with, for any $(i,j) \in [d_1] \times [d_2]$, according to definition of $\mathbf{S}_{ij}$, we have

$$\|\mathbf{S}_{ij}\|_2 \leq \frac{1}{p}\|\text{vec}(\mathcal{P}_\mathcal{T}(\mathbf{X}_{i,*}^\top \mathbf{Y}_{j,*}))\|_2^2 = \frac{1}{p}\|\mathcal{P}_\mathcal{T}(\mathbf{X}_{i,*}^\top \mathbf{Y}_{j,*})\|_F^2 \leq \frac{\mu_0 \mu_1 r(n_1 + n_2)}{pd_1 d_2}, \quad (\text{C.4})$$

where the first inequality holds because $|\langle \mathbf{A}, \mathbf{B} \rangle| \leq \|\mathbf{A}\|_F \cdot \|\mathbf{B}\|_F$, and the second inequality follows from (C.3). Similarly, for any $\mathbf{Z} \in \mathbb{R}^{n_1 \times n_2}$, we have

$$\Big\|\sum_{i,j} \mathbb{E}[\mathbf{S}_{ij}^2]\Big\|_2 = \frac{1-p}{p}\Big\|\sum_{i,j} \text{vec}(\mathcal{P}_\mathcal{T}(\mathbf{X}_{i,*}^\top \mathbf{Y}_{j,*})) \cdot \|\mathcal{P}_\mathcal{T}(\mathbf{X}_{i,*}^\top \mathbf{Y}_{j,*})\|_F^2 \cdot \text{vec}(\mathcal{P}_\mathcal{T}(\mathbf{X}_{i,*}^\top \mathbf{Y}_{j,*}))^\top\Big\|_2$$
$$\leq \frac{\mu_0 \mu_1 r(n_1 + n_2)}{pd_1 d_2} \cdot \sup_{\|\mathbf{Z}\|_F=1} \Big\|\sum_{i,j} \text{vec}(\mathcal{P}_\mathcal{T}(\mathbf{X}_{i,*}^\top \mathbf{Y}_{j,*}))\text{vec}(\mathcal{P}_\mathcal{T}(\mathbf{X}_{i,*}^\top \mathbf{Y}_{j,*}))^\top \text{vec}(\mathbf{Z})\Big\|_2$$
$$= \frac{\mu_0 \mu_1 r(n_1 + n_2)}{pd_1 d_2} \cdot \sup_{\|\mathbf{Z}\|_F=1} \Big\|\mathcal{P}_\mathcal{T}\Big(\sum_{i,j} \langle \mathcal{P}_\mathcal{T}(\mathbf{Z}), \mathbf{X}_{i,*}^\top \mathbf{Y}_{j,*}\rangle \mathbf{X}_{i,*}^\top \mathbf{Y}_{j,*}\Big)\Big\|_F$$
$$\leq \frac{\mu_0 \mu_1 r(n_1 + n_2)}{pd_1 d_2} \cdot \sup_{\|\mathbf{Z}\|_F=1} \big\|\mathcal{P}_\mathcal{T}\big(\mathbf{X}^\top \mathbf{X}\mathcal{P}_\mathcal{T}(\mathbf{Z})\mathbf{Y}^\top \mathbf{Y}\big)\big\|_F \leq \frac{\mu_0 \mu_1 r(n_1 + n_2)}{pd_1 d_2},$$

where the first equality follows from the definition of $\mathbf{S}_{ij}$, the first inequality follows from (C.3) and the definition of spectral norm, the second equality holds because $\langle \mathcal{P}_\mathcal{T}(\mathbf{A}), \mathbf{B}\rangle = \langle \mathbf{A}, \mathcal{P}_\mathcal{T}(\mathbf{B})\rangle$, and the second inequality holds because $\mathbf{X}, \mathbf{Y}$ are orthonormal and the projection operator $\mathcal{P}_\mathcal{T}$ is non-expansive. Thus, we obtain

$$\left\|\mathbb{E}(\mathbf{S}_{ij}^2)\right\|_2 \leq \frac{\mu_0\mu_1 r(n_1+n_2)}{pd_1d_2}. \tag{C.5}$$

Therefore, combining (C.4) and (C.5), according to Lemma D.1, for any $\gamma \in (0,1)$, we have

$$\mathbb{P}\left\{\left\|\sum_{i,j}\mathbf{S}_{ij}\right\|_2 \geq \gamma\right\} \leq (n_1+n_2)\cdot\exp\left(\frac{-\gamma^2/2}{(1+\gamma/3)\mu_0\mu_1 r(n_1+n_2)/(pd_1d_2)}\right) \leq c'/d,$$

under condition $p \geq c\mu_0\mu_1 rn\log(d)/(\gamma^2 d_1 d_2)$, where $c$ is a constant. Note that for all $\mathbf{Z} \in \mathbb{R}^{d_1\times d_2}$, we have

$$\left\|\sum_{i,j}\mathcal{S}_{ij}(\mathbf{Z})\right\|_F = \left\|\sum_{i,j}\mathbf{S}_{ij}\cdot\text{vec}(\mathbf{Z})\right\|_F \leq \left\|\sum_{i,j}\mathbf{S}_{ij}\right\|_2\cdot\|\mathbf{Z}\|_F,$$

we complete the proof of (B.2). Furthermore, for all $\mathbf{Z}_1, \mathbf{Z}_2 \in \mathcal{T}$, we have

$$\begin{aligned}\left|\langle\mathbf{X}\mathbf{Z}_1\mathbf{Y}^\top - p^{-1}\mathcal{P}_\Omega(\mathbf{X}\mathbf{Z}_1\mathbf{Y}^\top), \mathbf{X}\mathbf{Z}_2\mathbf{Y}^\top\rangle\right| &= \left|\left\langle\mathcal{P}_\mathcal{T}(\mathbf{Z}_1) - p^{-1}\mathcal{P}_\mathcal{T}\left(\mathbf{X}^\top\mathcal{P}_\mathcal{T}(\mathbf{X}\mathcal{P}_\mathcal{T}(\mathbf{Z}_1)\mathbf{Y}^\top)\mathbf{Y}^\top\right), \mathbf{Z}_2\right\rangle\right| \\ &\leq \left\|\mathcal{P}_\mathcal{T}(\mathbf{Z}_1) - p^{-1}\mathcal{P}_\mathcal{T}\left(\mathbf{X}^\top\mathcal{P}_\mathcal{T}(\mathbf{X}\mathcal{P}_\mathcal{T}(\mathbf{Z}_1)\mathbf{Y}^\top)\mathbf{Y}^\top\right)\right\|_F\cdot\|\mathbf{Z}_2\|_F \\ &\leq \gamma\|\mathbf{Z}_1\|_F\cdot\|\mathbf{Z}_2\|_F,\end{aligned}$$

where the first equality holds because $\langle\mathbf{A}, \mathcal{P}_\mathcal{T}(\mathbf{B})\rangle = \langle\mathcal{P}_\mathcal{T}(\mathbf{A}), \mathbf{B}\rangle$, the second inequality holds because $\langle\mathbf{A}, \mathbf{B}\rangle \leq \|\mathbf{A}\|_F\cdot\|\mathbf{B}\|_F$, and the last inequality follows from (B.2). Finally, for all $\mathbf{Z} \in \mathcal{T}$, we have

$$p^{-1}\left\|\mathcal{P}_\Omega(\mathbf{X}\mathbf{Z}\mathbf{Y}^\top)\right\|_F^2 \leq \left|\langle\mathbf{Z} - p^{-1}\mathbf{X}^\top\mathcal{P}_\Omega(\mathbf{X}\mathbf{Z}\mathbf{Y}^\top)\mathbf{Y}, \mathbf{Z}\rangle\right| + \|\mathbf{Z}\|_F^2 \leq (1+\gamma)\|\mathbf{Z}\|_F^2,$$

which complete the proof. $\square$

### C.3 Proof of Lemma B.3

*Proof.* Note that we have

$$\begin{aligned}\frac{1}{p}\left\|\mathcal{P}_\Omega(\mathbf{X}_L\mathbf{U}\mathbf{V}^\top\mathbf{X}_R^\top)\right\|_F^2 &= \|\mathbf{X}_L\mathbf{U}\mathbf{V}^\top\mathbf{X}_R^\top\|_F^2 + \langle p^{-1}\mathcal{P}_\Omega(\mathbf{X}_L\mathbf{U}\mathbf{V}^\top\mathbf{X}_R^\top) - \mathbf{X}_L\mathbf{U}\mathbf{V}\mathbf{X}_R^\top, \mathbf{X}_L\mathbf{U}\mathbf{V}^\top\mathbf{X}_R^\top\rangle \\ &= \|\mathbf{U}\mathbf{V}^\top\|_F^2 + \langle(p^{-1}\mathcal{P}_\Omega - \mathcal{I})(\mathbf{X}_L\mathbf{U}\mathbf{V}^\top\mathbf{X}_R^\top), \mathbf{X}_L\mathbf{U}\mathbf{V}^\top\mathbf{X}_R^\top\rangle \\ &= \|\mathbf{U}\mathbf{V}^\top\|_F^2 + \sum_{(i,j)=(1,1)}^{(d_1,d_2)}\left(\frac{\xi_{ij}}{p} - 1\right)\cdot\left[\mathbf{X}_L\mathbf{U}\mathbf{V}^\top\mathbf{X}_R^\top\right]_{ij}^2,\end{aligned}$$

where $\xi_{ij} = 1$, if $(i,j) \in \Omega$; $\xi_{ij} = 0$, otherwise. Thus, it is sufficient to bound the second term on the right hand side. For simplicity, we let $\alpha_{ij} = (p^{-1}\xi_{ij} - 1)\cdot[\mathbf{X}_L\mathbf{U}\mathbf{V}^\top\mathbf{X}_R^\top]_{ij}^2$. Note that $\mathbf{U}, \mathbf{V}$ are fixed, we have $\mathbb{E}[\alpha_{ij}] = 0$. In addition, we can upper bound $|\alpha_{ij}|$ by

$$\begin{aligned}|\alpha_{ij}| &\leq \frac{1}{p}\|\mathbf{X}_L\mathbf{U}\mathbf{V}^\top\mathbf{X}_R^\top\|_{\infty,\infty}^2 \leq \frac{1}{p}\max\left\{\|\mathbf{X}_L\|_{2,\infty}^2\cdot\|\mathbf{U}\|_F^2\cdot\|\mathbf{X}_R\mathbf{V}\|_{2,\infty}^2, \|\mathbf{X}_L\mathbf{U}\|_{2,\infty}^2\cdot\|\mathbf{V}\|_F^2\cdot\|\mathbf{X}_R\|_{2,\infty}^2\right\} \\ &\leq \frac{9\mu_0\mu_1 r\sigma_1^*(n_1+n_2)}{pd_1d_2}\cdot(\|\mathbf{U}\|_F^2 + \|\mathbf{V}\|_F^2),\end{aligned}$$

where the second inequality follows from Assumption 3.2. Next, we are going to bound the variance

$$\text{Var}\Big(\sum_{(i,j)=(1,1)}^{(d_1,d_2)} \alpha_{ij}\Big) = \sum_{(i,j)=(1,1)}^{(i,j)} \text{Var}(\alpha_{ij})$$

$$\leq \frac{1}{p} \sum_{(i,j)=(1,1)}^{(d_1,d_2)} \big[\mathbf{X}_L \mathbf{U} \mathbf{V}^\top \mathbf{X}_R^\top\big]_{ij}^4$$

$$\leq \frac{1}{p}\|\mathbf{X}_L \mathbf{U} \mathbf{V}^\top \mathbf{X}_R^\top\|_{\infty,\infty}^2 \cdot \|\mathbf{X}_L \mathbf{U} \mathbf{V}^\top \mathbf{X}_R^\top\|_F^2 \leq \frac{9\mu_0^2 r^2 \sigma_1^{*2}}{pd_1 d_2}(\|\mathbf{U}\|_F^2 + \|\mathbf{V}\|_F^2)^2,$$

where equality holds because $\alpha_{ij}$'s are independent, and the last inequality follows from the assumptions $\|\mathbf{X}_L \mathbf{U}\|_{2,\infty} \leq 3\sqrt{\mu_0 r \sigma_1^*/d_1}$, $\|\mathbf{X}_R \mathbf{V}\|_{2,\infty} \leq 3\sqrt{\mu_0 r \sigma_1^*/d_2}$. Therefore, applying bernstein inequality for random variables, under condition that $|\Omega| \geq c \max\{\mu_0^2 r^2 \kappa^2, \mu_0 \mu_1 r \kappa n\} \log d/\gamma^2$, with probability at least $1 - c'/d$, we have

$$\Big|\sum_{(i,j)=(1,1)}^{(d_1,d_2)} \alpha_{ij}\Big| \leq \gamma \sigma_r^* \cdot (\|\mathbf{U}\|_F^2 + \|\mathbf{V}\|_F^2),$$

which completes the proof. $\square$

## C.4 Proof of Lemma B.4

*Proof.* For any $i \in [d_1]$, we denote $\mathbf{X}_{i,*} \in \mathbb{R}^{1 \times n_1}$ as the $i$-th row vector of $\mathbf{X}$. Similarly, for any $j \in [d_2]$, we denote $\mathbf{Y}_{j,*} \in \mathbb{R}^{1 \times n_2}$ as the $j$-th row vector of $\mathbf{Y}$. Besides, for any $(i,j) \in [d_1] \times [d_2]$, let $\xi_{ij} = 1$, if $(i,j) \in \Omega$, and zero otherwise. Note that $\mathbf{X}, \mathbf{Y}$ are orthonormal, thus for any $\mathbf{Z} \in \mathbb{R}^{n_1 \times n_2}$, we have

$$\mathbf{Z} - p^{-1}\mathbf{X}^\top \mathcal{P}_\Omega\big(\mathbf{X}\mathbf{Z}\mathbf{Y}^\top\big)\mathbf{Y} = \mathbf{X}^\top(\mathcal{I} - p^{-1}\mathcal{P}_\Omega)(\mathbf{X}\mathbf{Z}\mathbf{Y}^\top)\mathbf{Y}$$

$$= \sum_{(i,j)\in[d_1]\times[d_2]} (1 - p^{-1}\xi_{ij})\langle \mathbf{X}\mathbf{Z}\mathbf{Y}^\top, \mathbf{e}_i \mathbf{e}_j^\top \rangle \mathbf{X}^\top \mathbf{e}_i \mathbf{e}_j^\top \mathbf{Y}$$

$$= \sum_{(i,j)\in[d_1]\times[d_2]} (1 - p^{-1}\xi_{ij})\langle \mathbf{Z}, \mathbf{X}_{i,*}^\top \mathbf{Y}_{j,*} \rangle \mathbf{X}_{i,*}^\top \mathbf{Y}_{j,*},$$

where $\mathcal{I} : \mathbb{R}^{d_1 \times d_2} \to \mathbb{R}^{d_1 \times d_2}$ is the identity mapping, and $\mathbf{e}_i$ denotes the $i$-th standard basis. For any $(i,j) \in [d_1] \times [d_2]$, define linear operator $\mathcal{S}_{ij} : \mathbb{R}^{n_1 \times n_2} \to \mathbb{R}^{n_1 \times n_2}$, such that

$$\mathcal{S}_{ij}(\mathbf{Z}) = (1 - p^{-1}\xi_{ij})\langle \mathbf{Z}, \mathbf{X}_{i,*}^\top \mathbf{Y}_{j,*} \rangle \mathbf{X}_{i,*}^\top \mathbf{Y}_{j,*}.$$

Define $\mathbf{S}_{ij} \in \mathbb{R}^{d_1 d_2 \times d_1 d_2}$ as the corresponding matrix to the linear operator $\mathcal{S}_{ij}$ such that

$$\mathbf{S}_{ij} = (1 - p^{-1}\xi_{ij})\text{vec}(\mathbf{X}_{i,*}^\top \mathbf{Y}_{j,*})\text{vec}(\mathbf{X}_{i,*}^\top \mathbf{Y}_{j,*})^\top, \tag{C.6}$$

then we can easily show that $\|\sum_{i,j} \mathcal{S}_{ij}(\mathbf{Z})\|_F^2 = \|\sum_{i,j} \mathbf{S}_{ij} \cdot \text{vec}(\mathbf{Z})\|_2^2$, for any $\mathbf{Z} \in \mathbb{R}^{d_1 \times d_2}$. Obviously, we have $\mathbb{E}[\mathbf{S}_{ij}] = \mathbf{0}$ and $\mathbf{S}_{ij}$ is symmetric. For any $(i,j) \in [d_1] \times [d_2]$, according to the definition of $\mathbf{S}_{ij}$ in (C.6), we have

$$\|\mathbf{S}_{ij}\|_2 \leq \frac{1}{p}\|\text{vec}(\mathbf{X}_{i,*}^\top \mathbf{Y}_{j,*})\|_2^2 = \frac{1}{p}\|\mathbf{X}_{i,*}^\top \mathbf{Y}_{j,*}\|_F^2 \leq \frac{1}{p}\|\mathbf{X}\|_{2,\infty}^2 \cdot \|\mathbf{Y}\|_{2,\infty}^2 \leq \frac{\mu_1^2 n_1 n_2}{pd_1 d_2}, \tag{C.7}$$

where the second inequality holds because $\|\mathbf{X}_{i,*}^\top \mathbf{Y}_{j,*}\|_F = \|\mathbf{X}_{i,*}\|_2 \cdot \|\mathbf{Y}_{j,*}\|_2$, and the last inequality follows from the fact that $\mathbf{X}, \mathbf{Y}$ are $\mu_1$ self-incoherent. Moreover, for any $\mathbf{Z} \in \mathbb{R}^{n_1 \times n_2}$, we have

$$\begin{aligned}
\Big\|\sum_{i,j} \mathbb{E}[\mathbf{S}_{ij}^2]\Big\|_2 &= \frac{1-p}{p} \Big\|\sum_{i,j} \text{vec}(\mathbf{X}_{i,*}^\top \mathbf{Y}_{j,*}) \cdot \|\mathbf{X}_{i,*}^\top \mathbf{Y}_{j,*}\|_F^2 \cdot \text{vec}(\mathbf{X}_{i,*}^\top \mathbf{Y}_{j,*})^\top \Big\|_2 \\
&\leq \frac{\mu_1^2 n_1 n_2}{p d_1 d_2} \cdot \sup_{\|\mathbf{Z}\|_F = 1} \Big\|\sum_{i,j} \text{vec}(\mathbf{X}_{i,*}^\top \mathbf{Y}_{j,*}) \text{vec}(\mathbf{X}_{i,*}^\top \mathbf{Y}_{j,*})^\top \text{vec}(\mathbf{Z}) \Big\|_2 \\
&= \frac{\mu_1^2 n_1 n_2}{p d_1 d_2} \cdot \sup_{\|\mathbf{Z}\|_F = 1} \Big\|\sum_{i,j} \langle \mathbf{Z}, \mathbf{X}_{i,*}^\top \mathbf{Y}_{j,*} \rangle \mathbf{X}_{i,*}^\top \mathbf{Y}_{j,*} \Big\|_F \\
&= \frac{\mu_1^2 n_1 n_2}{p d_1 d_2} \cdot \sup_{\|\mathbf{Z}\|_F = 1} \|\mathbf{X}^\top \mathbf{X} \mathbf{Z} \mathbf{Y}^\top \mathbf{Y}\|_F \leq \frac{\mu_1^2 n_1 n_2}{p d_1 d_2},
\end{aligned}$$

where the first equality follows from the definition of $\mathbf{S}_{ij}$ in (C.6), the first inequality follows from the fact that $\|\mathbf{X}_{i,*}^\top \mathbf{Y}_{j,*}\|_F^2 \leq \mu^2 n_1 n_2/(d_1 d_2)$, and the last equality holds because $\mathbf{X}, \mathbf{Y}$ are both orthonormal. Therefore, according to matrix bernstein inequality as in Lemma D.1, for any $\gamma \in (0, 1)$, we have

$$\mathbb{P}\Big\{\Big\|\sum_{i,j} \mathbf{S}_{ij}\Big\|_2 \geq \mu_1 n \gamma \Big\} \leq 2 n_1 n_2 \cdot \exp\Big(\frac{-\gamma^2 \mu_1^2 n^2 / 2}{(1 + \gamma \mu_1 n / 3) \mu_1^2 n_1 n_2 / (p d_1 d_2)}\Big) \leq c'/d,$$

under condition $|\Omega| \geq c \mu_1 n \log d / \gamma^2$, where $c$ is a constant. Thus according to the definition of $\mathcal{S}_{ij}$, for all $\mathbf{Z} \in \mathbb{R}^{n_1 \times n_2}$, with probability at least $1 - c'/d$, we have

$$\|\mathbf{Z} - p^{-1} \mathbf{X}^\top \mathcal{P}_\Omega(\mathbf{X} \mathbf{Z} \mathbf{Y}^\top) \mathbf{Y}\|_F = \Big\|\sum_{i,j} \mathcal{S}_{ij}(\mathbf{Z})\Big\|_F = \Big\|\sum_{i,j} \mathbf{S}_{ij} \cdot \text{vec}(\mathbf{Z})\Big\|_2 \leq \Big\|\sum_{i,j} \mathbf{S}_{ij}\Big\|_2 \cdot \|\mathbf{Z}\|_F \leq \mu_1 n \gamma \|\mathbf{Z}\|_F.$$

By triangle's inequality, we complete the proof. □

# D  Additional Auxiliary Lemmas

In this section, we provide the bernstein inequalities used in the proofs for our main results.

**Lemma D.1.** (Tropp, 2012) Consider a finite sequence $\{\mathbf{Z}_k\}$ of independent random matrices with dimension $d_1 \times d_2$. Assume that each random matrix satisfies

$$\mathbb{E}(\mathbf{Z}_k) = \mathbf{0} \quad \text{and} \quad \|\mathbf{Z}_k\|_2 \leq R \quad \text{almost surely.}$$

Define

$$\sigma^2 = \max\Big\{\Big\|\sum_k \mathbb{E}(\mathbf{Z}_k \mathbf{Z}_k^\top)\Big\|_2, \Big\|\sum_k \mathbb{E}(\mathbf{Z}_k^\top \mathbf{Z}_k)\Big\|_2\Big\}.$$

Then, for all $t \geq 0$, we have

$$\mathbb{P}\Big\{\Big\|\sum_k \mathbf{Z}_k\Big\|_2 \geq t\Big\} \leq (d_1 + d_2) \cdot \exp\Big(\frac{-t^2/2}{\sigma^2 + Rt/3}\Big).$$

**Lemma D.2.** Assume the index set $\Omega$ follows Bernoulli model (3.1). There exists constants $c, c'$ such that under condition that $|\Omega| \geq c \log d$, with probability at least $1 - c'/d$, we have

$$\big||\Omega| - p d_1 d_2\big| \leq \frac{1}{2} p d_1 d_2,$$

where $p = |\Omega|/(d_1 d_2)$.

Lemma D.2 can be directly derived from the bernstein-type inequality for independent random variables.